\documentclass[10pt,twocolumn,letterpaper]{article}

\usepackage{cvpr}              

\definecolor{cvprblue}{rgb}{0.21,0.49,0.74}
\usepackage[pagebackref,breaklinks,colorlinks,allcolors=cvprblue]{hyperref}

\usepackage{bm}
\usepackage{multirow}
\usepackage{verbatim}
\usepackage{multirow}
\usepackage{graphicx}
\usepackage[normalem]{ulem}
\useunder{\uline}{\ul}{}
\usepackage[misc]{ifsym} 

\usepackage[capitalize]{cleveref}

\def\paperID{5642} 
\def\confName{CVPR}
\def\confYear{2025}

\title{Visual-Instructed Degradation Diffusion for All-in-One Image Restoration}

\author{Wenyang Luo\textsuperscript{1,2*}, Haina Qin\textsuperscript{1,2*$\heartsuit$}, Zewen Chen\textsuperscript{1,2*}, Libin Wang\textsuperscript{3$\dagger$},\\ Dandan Zheng\textsuperscript{3}, Yuming Li\textsuperscript{3}, Yufan Liu\textsuperscript{1,2}, Bing Li\textsuperscript{1,2,4,$\textrm{\Letter}$} , Weiming Hu\textsuperscript{1,2} \\ 
\textsuperscript{1}State Key Laboratory of Multimodal Artificial Intelligence Systems (MAIS), CASIA;\\
\textsuperscript{2}School of Artificial Intelligence, University of Chinese Academy of Sciences;\ \ \textsuperscript{3}Ant Group;\\\textsuperscript{4}PeopleAI Inc.\\
{\tt\small \{luowenyang2020@,qinhaina2020@,bli@nlpr.,wmhu@nlpr.\}ia.ac.cn lbin.wlb@antgroup.com}
}

\begin{document}
\maketitle
{\renewcommand{\thefootnote}{*}\footnotetext{Equal contribution. \href{https://github.com/luowyang/Defusion}{Project Page}. $\textrm{\Letter}$ Corresponding author.}}
{\renewcommand{\thefootnote}{$\heartsuit$}\footnotetext{Work done during internship at Ant Group. $\dagger$ Project lead.}}

\begin{abstract}
Image restoration tasks like deblurring, denoising, and dehazing usually need distinct models for each degradation type, restricting their generalization in real-world scenarios with mixed or unknown degradations. In this work, we propose \textbf{Defusion}, a novel all-in-one image restoration framework that utilizes visual instruction-guided degradation diffusion. Unlike existing methods that rely on task-specific models or ambiguous text-based priors, Defusion constructs explicit \textbf{visual instructions} that align with the visual degradation patterns. These instructions are grounded by applying degradations to standardized visual elements, capturing intrinsic degradation features while agnostic to image semantics. Defusion then uses these visual instructions to guide a diffusion-based model that operates directly in the degradation space, where it reconstructs high-quality images by denoising the degradation effects with enhanced stability and generalizability. Comprehensive experiments demonstrate that Defusion outperforms state-of-the-art methods across diverse image restoration tasks, including complex and real-world degradations.
\end{abstract}


    
\vspace{-0.4cm}
\section{Introduction}
\label{sec:intro}



Image restoration is a fundamental task in computer vision, aiming to recover high-quality images from their degraded counterparts. This encompasses various challenges, including denoising \cite{anwar2019real, wang2022blind2unblind, ren2021adaptive}, deblurring \cite{suin2020spatially, whang2022deblurring, tsai2022stripformer}, dehazing \cite{liu2019griddehazenet, song2023vision}, deraining \cite{jiang2020multi, chen2023learning, chen2023sparse}, JPEG artifact removal \cite{jiang2021towards, ehrlich2020quantization}, etc. Traditional deep learning-based approaches have achieved remarkable progress by developing task-specific models that excel in individual restoration tasks. However, these approaches often struggle with generalization, requiring separate models for each degradation type, which limits their practical applicability in real-world scenarios with complex degradations.

\begin{figure}[t]
    \centering
    \includegraphics[width=\linewidth]{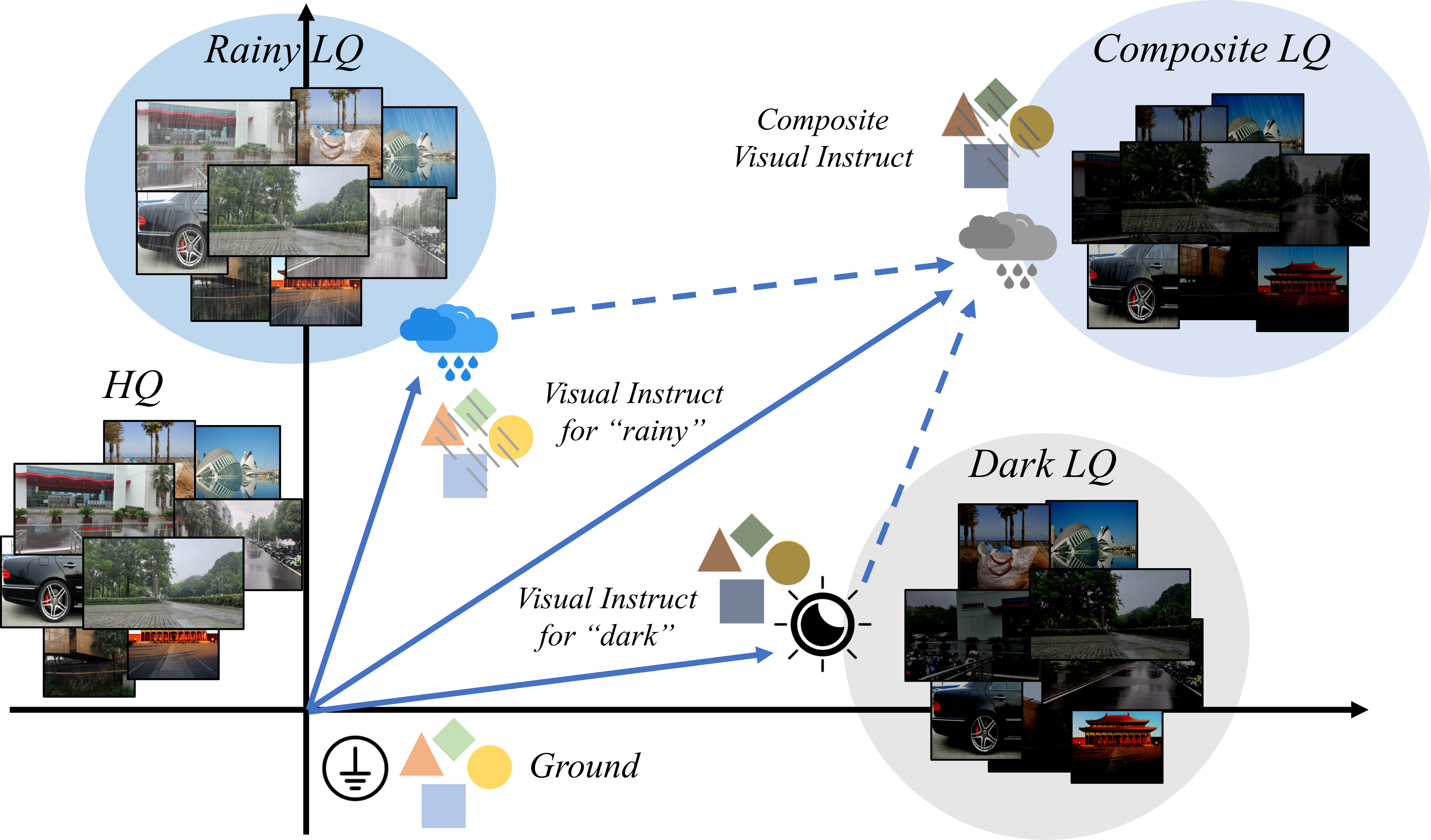}
    \caption{Visual instructions speak for visual degradations. Visual instructions are constructed by applying the artifacts of image degradations upon the composition of basic image elements (dubbed visual grounds). They faithfully demonstrate the visual effects of degradations and their compositions.}
    \label{fig:intro}
    \vspace{-0.4cm}
\end{figure}

To address this limitation, all-in-one image restoration methods handling various degradations in a single model have gained significant attention \cite{lin2023diffbir, zhang2024perceive, liu2024diff, ai2024multimodal, potlapalli2024promptir, park2023all}. They use implicit or explicit prior knowledge to identify degradation patterns and guide restoration. However, existing methods face challenges in deriving effective priors that align with the visual structures of degraded images.

Implicit priors are extracted from input low-quality (LQ) images by task-specific sub-networks \cite{chen2021pre, park2023all} or learnable prompts \cite{potlapalli2024promptir, lin2023diffbir, zhang2024perceive, ai2024multimodal}; they are essentially equivalent to increasing the parameter counts, limiting their generalizability to complex degradations \cite{jiang2023autodir}. 
Explicit priors, typically text descriptions processed by language models \cite{conde2024instructir, yu2024scaling}, suffer from weak alignment with low-level visual structures \cite{wu2023q, liu2025mmbench}. Text descriptions are typically coarse or agnostic to degradations \cite{chen2024lion} due to inherent language ambiguity and a lossy language-vision bridge. Thus, the text descriptions may collapse into category labels, crippling the benefit of the added complexity.

Given the above limitations, we propose to \textit{let visual speak for visual} and guide the image restoration model with \textbf{visual instructions} as explicit priors. Our visual instructions are constructed to demonstrate the artifacts of image degradations as visual guidance. This is achieved by applying the image corruption process (e.g. raining or noising) of the LQ images upon the visual grounds consisting of basic visual elements such as random textures, basic shapes, standard colors, etc, as shown in \cref{fig:intro}. The resulting visual instructions encode the exact visual effects (e.g. rain lines or noisy ``dots'') of the degradation with a minimal loss of information. The primary benefit is that the resulting visual instructions are inherently aligned with visual degradation patterns. For example, the combined visual instruction in \cref{fig:intro} can naturally show similar distortions as the composite LQ images. This makes the visual instructions suitable for low-level vision tasks.

We propose a unified diffusion-based image restoration framework, \underline{\textbf{De}}gradation Dif\underline{\textbf{fusion}}, which is guided by our visual instructions to denoise in the degradation space.
The diffusion model provides generative priors for the restored images, conditioned on the LQ images through an Image Restoration Bridge (IRB) to preserve the structural priors of the LQ images and maintain visual consistency between LQ and restored images.
Visual instructions, which capture the prominent intrinsic artifacts that degradations induce on textures, colors, lightness, and other image elements, control the model to perform restoration tasks. Specifically, visual instructions are ``tokenized'' and hint the model of the degradations in the LQ images via a Visual Instruct Adapter (VIA).
Moreover, Defusion operates within the degradation space defined as the space of transformations between LQ and HQ images, which aligns with the visual instructions describing the degradation.
The restoration model is jointly trained on multiple tasks. During inference, it follows visual instructions to handle various degradation types.

Our contributions are summarized as follows:
\begin{enumerate}
    \item We introduce a novel diffusion-based all-in-one image restoration framework, dubbed Defusion, guided by visual instructions to denoise in the degradation space.
    \item Our visual instructions visually align with the image degradations and retain their visual compositional properties, giving them advantages in low-level vision tasks.
    \item We conducted comprehensive experiments on various image restoration tasks under different scenarios including complex and unseen degradations, and Defusion consistently outperforms state-of-the-art methods.
\end{enumerate}

\section{Related Work}
\label{sec:related}

\subsection{All-in-One Image Restoration}

Image restoration methods mainly target predefined image degradation \cite{zheng2023empowering, Lin2023UnsupervisedID, Jin2023DNFDA, Li2023EmbeddingFF, Guo2022ImageDT, Fang2022ARN, liang2021swinir, li2022learning, chen2022simple, zamir2022restormer} such as motion blurring or rainy images. Such methods require developing and deploying separate models for each kind of degradation.
Recently, several all-in-one approaches \cite{qin2024restore, conde2024instructir, guo2024onerestore, jiang2023autodir, li2024promptcir, lin2023diffbir, zhang2024perceive, yu2024scaling, ye2024learning, liu2024diff, ai2024multimodal, potlapalli2024promptir, ma2023prores, zhang2023real, park2023all, zheng2024selective} have been proposed to tackle multiple degradations with one unified model. All-in-one approaches have the benefit that the model can learn the common knowledge from mixed degradations and possibly generalize to composite or unseen degradation.

Most all-in-one approaches \cite{qin2024restore, guo2024onerestore, li2024promptcir, lin2023diffbir, zhang2024perceive, ye2024learning, liu2024diff, ai2024multimodal, potlapalli2024promptir, ma2023prores, zhang2023real, park2023all, zheng2024selective} adopt auxiliary modules or learnable prompts to detect degradation types for the restoration model. These designs are often equivalent to increasing model parameter count, which hinders their generalizability to complex degradations.

SUPIR \cite{yu2024scaling}, InstructIR \cite{conde2024instructir}, and AutoDIR \cite{jiang2023autodir} guide image restoration with text-based instructions and utilize the concepts the pretrained language models learned to generalize to multiple degradations. However, language is essentially ambiguous and weakly aligned with visual features. For example, ``enhance the image'' can either mean ``increase lightness'' or ``denoising''; ``underwater image enhancement'' can be visually decomposed into ``lightness enhancement'' and ``deblurring'', which is not obvious from language.

In contrast, we propose accurately guiding image restoration with visual instructions, which closely align with the underlying visual degradations. Our visual instructions directly reveal distorted images' visual artifacts. The restoration model can learn the individual distortion patterns and their combinations during training and generalize to new degradations with the hints of our visual instructions.

\begin{figure*}[ht]
    \centering
    \includegraphics[width=\textwidth]{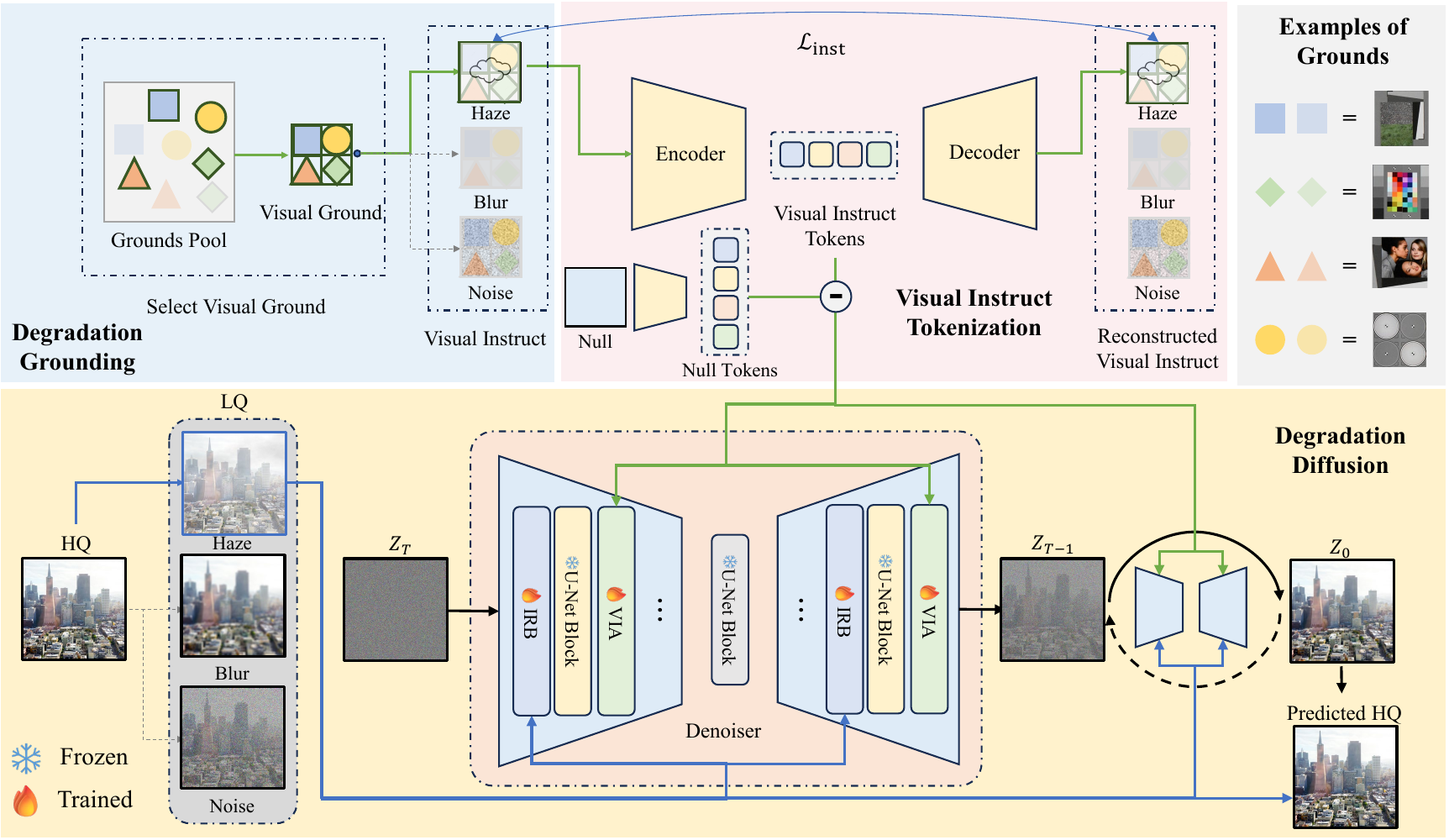}
    \caption{Framework of the proposed method. 1) Visual instructions are constructed from visual grounds to demonstrate the visual effects of the image degradations, 2) visual instructions are tokenized and contrasted with the ``clean'' (null) visual grounds, 3) finally, the visual instruction tokens guide the denoising diffusion model which estimates the degradation according to the hint of visual instructions. The visual instruction tokenizer is a quantized auto-encoder trained by the loss $\mathcal{L}_\text{inst}$, while only the encoder is used for inference.}
    \label{fig:framework}
\end{figure*}

\subsection{Diffusion Models for Image Restoration}

Image restoration can be formulated as an image-to-image translation task, with low-quality (LQ) images as input and high-quality (HQ) images as output. Diffusion-based generative models (DMs) \cite{sohl2015deep, ho2020denoising, song2020score} have risen as attractive solutions due to their remarkable ability to produce natural and visually appealing HQ images \cite{dhariwal2021diffusion, rombach2022high, luo2023refusion, kawar2022denoising, ding2024restoration}.

Many works \cite{luo2023refusion, xia2023diffir, ai2024multimodal, liu2024diff, yu2024scaling, lin2023diffbir, jiang2023autodir, zhang2024diffusion} adopt the latent diffusion architecture following Stable Diffusion \cite{rombach2022high}, where a separate autoencoder (VAE) \cite{kingma2013auto} extracts latent codes from the LQ images to denoise. However, the VAE discards most low-level details, which harms image restoration.

Other works operate on pixels \cite{liu2024structure, kawar2022denoising, luo2023image} or frequency domain \cite{zhao2024wavelet}.
Among them, some works propose dedicated diffusion processes to model image degradations, where the HQ images are corrupted given the LQ images \cite{kawar2022denoising} or shifts between HQ and LQ images \cite{luo2023image, liu2024structure, liu2024residual, shi2023resfusion, yue2024resshift}. These complicated approaches usually make approximate assumptions about HQ/LQ images or their shifts, leading to increased accumulated errors during inference. Moreover, they require training from scratch, unable to leverage prior knowledge about natural (clean) HQ images in pretrained models \cite{rombach2022high, saharia2022photorealistic, ramesh2022hierarchical}.
Instead, we diffuse the degradations between HQ and LQ conditioned on LQ images, guided by our visual instructions via the Visual Instruction Adapter (VIA). Our approach is simple yet effective for all-in-one image restoration, allowing for stable training and easy adaptation from pretrained models.

\subsection{Visual Instructions in Diffusion Models}

Previous works guide diffusion models to generate images with various types of instructions, including both texts \cite{rombach2022high, saharia2022photorealistic, ramesh2022hierarchical, feng2024layoutgpt, lian2023llm} and visual instructions such as depth maps \cite{zhang2023adding}, segmentation maps \cite{couairon2022diffedit, wang2022pretraining}, layouts \cite{rombach2022high}, inpainting masks \cite{huang2023composer, xie2023smartbrush}, sketches \cite{voynov2023sketch}, skeletons \cite{mou2024t2i}, keypoints \cite{li2023gligen}, example images \cite{bar2022visual, wang2023images, nguyen2024visual}, to name a few.
Visual instructions are constructed to demonstrate image generation requirements or desired edits that are difficult to describe precisely in languages, such as turning the image into a drawing of a specific style \cite{nguyen2024visual}. In our work, we develop a novel visual instruction dedicated to image restoration, which more accurately captures the subtle low-level visual details of degradations than texts such as ``noise'' or ``blurring'', leading to better performance.

\section{Method}
\label{sec:method}

This section introduces the proposed diffusion-based all-in-one image restoration method, Defusion. \cref{fig:framework} depicts the overall framework. 
\cref{subsec:visual_instructions} describes the construction of our visual instructions.
\cref{subsec:degradation_diffusion} formulates the instruction-guided diffusion model in degradation space. \cref{subsec:conditioning} details the conditioning mechanisms with the visual instructions via a Visual Instruction Adapter (VIA) and condition generation on the LQ images with an Image Restoration Bridge (IRB).

\subsection{Visual Instructions}
\label{subsec:visual_instructions}

\noindent
\textbf{Motivation.}
Most all-in-one image restoration estimate degradation types given the input LQ images to guide restoration \cite{lin2023diffbir, ai2024multimodal, potlapalli2024promptir, park2023all}, where the estimation becomes inaccurate for unseen degradation types.
Inspired by InstructPix2Pix \cite{brooks2023instructpix2pix}, some work \cite{conde2024instructir, jiang2023autodir} control the restoration process with user-provided text instructions such as ``make this blurry images sharper''.
However, \textit{languages and visual patterns reside in very different spaces}. While the high-level semantics can be aligned by visual-language pretraining \cite{radford2021learning, alayrac2022flamingo, liu2024visual}, language cannot accurately convey visual details \cite{monsefi2024detailclip, shao2023investigating} such as how the degradations rearrange and combine pixels.

On the contrary, we \textit{let visual speak for visual} and resort to accurately describing the visual effect of degradations through visual instructions. Ideally, the visual instructions should capture the visually relevant aspects of degradations with minimal information loss and be disentangled with image semantics to improve generalizability. Next, we describe how to construct such visual instructions.

\vspace{0.5em}
\noindent
\textbf{Grounding of Degradations.}
Image degradations are ``dangling,'' meaning that their visual effects only manifest when they exist in the context of degraded images. Therefore, we first apply degradations on some ``standard images'' to visually demonstrate the degradations to the restoration model. We call this process \textit{grounding} of degradations, the ``standard images'' are dubbed visual grounds.

The visual grounds are carefully chosen to include various types of visual constructs, patterns, structures, etc., that may occur in natural images to reveal the full extent of degradation. We draw inspiration from image quality assessment \cite{mittal2012making, venkatanath2015blind, mittal2012no} and select TE42 \cite{te42}, a family of charts commonly used for camera testing and visual analysis, to construct a pool of visual grounds. TE42 comprises a rich combination of textures and color palettes. We group the constructs in TE42 charts into concentric textures, random textures, natural textures, and colors and randomly select one construct from each group to form a visual ground. \cref{fig:framework} shows some examples of visual grounds.
Given an arbitrary degradation, we transform the visual ground by the degradation with random parameters. The degraded visual ground forms the visual instruction.

\vspace{0.5em}
\noindent
\textbf{Degradation Tokenization.}
The grounded degradations (i.e. our visual instructions) obtained above can be regarded as ``visual words'' that ``speak'' how the degradations visually distort the image content.
We further ``tokenize'' the grounded degradations into \textit{degradation tokens} to guide diffusion-based image restoration, analogous to how text instructions are tokenized and guide text-to-image generation.

In our work, we tokenize our visual instructions with a Degradation Tokenizer that produces a sequence of discrete degradation token embeddings. Degradation Tokenizer compresses the perceptually unimportant constituents in the visual instructions with a quantization-regularized autoencoder inspired by LDM \cite{rombach2022high} and VQ-VAE \cite{van2017neural}.

Specifically, a visual instruction $\bm{v} \in \mathbb{R}^{H_v \times W_v \times 3}$ is mapped to embedding $\bm{e}_v \in \mathbb{R}^{h_v \times w_v \times C_v}$ ($C_V$ is latent dimension) by an encoder $\bm{E}$. Embedding $\bm{e}_v$ is then vector-quantized (VQ) into one of the \textit{degradation codes} $\bm{z}_v \in \mathbb{R}^{h_v \times w_v \times C_v}$ in the codebook $\mathcal{Z} = \{\bm{e}_k\}_{k=1}^K$ of size $K$ by
\begin{equation}
    \bm{z}_v
    = \arg\min_{\bm{e}_k\in\mathcal{Z}} \left\{ \| \bm{e}_k - \bm{e}_v \|_2
    = \| \bm{e}_k - \bm{E}(\bm{v}) \|_2 \right\},
\end{equation}
where $\|\cdot\|_2$ is Euclidean distance. The degradation code $\bm{z}_v$ is further reconstructed by a decoder $\bm{D}$ to approximate the original visual instruction $\bm{\hat{v}} = \bm{D}(\bm{z}_v)$.

\vspace{0.5em}
\noindent
\textbf{Training.}
The encoder $\bm{E}$, decoder $\bm{D}$ and degradation codebook $\mathcal{Z}$ are jointly optimized by the following loss:
\begin{align}
    &\mathcal{L}_\text{inst} = \mathcal{L}_\text{rec} + \mathcal{L}_\text{VQ}, \label{eq:loss_1} \\
    &\mathcal{L}_\text{VQ} = \|\text{sg}\left[\bm{E}(\bm{v}) - \bm{z}_v\right]\| + \|\bm{E}(\bm{v}) - \text{sg}\left[\bm{z}_v\right]\|, \label{eq:loss_2} \\
    &\mathcal{L}_\text{rec} = \|\bm{D}(\bm{z}_v) - \bm{v}\|_2^2 + \lambda\, \mathcal{L}_\text{LPIPS}(\bm{D}(\bm{z}_v), \bm{v}), \label{eq:loss_3}
\end{align}
where $\text{sg}[\cdot]$ is the stop-gradient operator \cite{van2017neural}, and the gradient back-propagates through $\bm{z}_v$ by straight-through estimator \cite{bengio2013estimating}. The reconstruction loss $\mathcal{L}_\text{rec}$ consists of a mean-squared error (MSE) and a perceptual loss $\mathcal{L}_\text{LPIPS}$ \cite{zhang2018unreasonable} to improve the perception of visual details, balanced by hyperparameter $\lambda$. To improve visual quality, we also adopt a hinge-based adversarial loss \cite{rombach2022high, esser2021taming}. We randomly replace the visual instruction $\bm{v}$ with its clean visual ground with a probability of $0.1$ during training.

\vspace{0.5em}
\noindent
\textbf{Inference.}
The trained encoder $\bm{E}$ serves as our Degradation Tokenizer. The visual instructions discretized and encoded by $\bm{E}$ capture the essential visual patterns and structures of degradations.
To improve the discriminability of degradations, we also encode the clean visual grounds (``null'' in \cref{fig:framework}) without degradation and subtract their embeddings from those of $\bm{v}$ and use the residuals to guide our degradation diffusion in \cref{subsec:degradation_diffusion}. 

\subsection{Degradation Diffusion}
\label{subsec:degradation_diffusion}

\noindent
\textbf{Preliminary: Diffusion.}
Diffusion models \cite{ho2020denoising, kingma2021variational} generate data $\bm{x}_0 \sim p_0(\cdot)$ by reversing the following forward process that ``diffuses'' $\bm{x}_0$ into standard Gaussian noise $\bm{\epsilon} \sim \mathcal{N}(\bm{0}, \bm{I})$:
\begin{equation}
    p_t(\bm{x}_t|\bm{x}_0) = \mathcal{N}(\alpha_t \bm{x}_0, \sigma_t^2 \bm{I}); \quad 0 \le t \le 1,
    \label{eq:diffusion}
\end{equation}
where $\alpha_t, \sigma_t$ define the noise schedule.
The reversing is achieved by solving the reverse-time stochastic differential equation (SDE) \cite{song2020score, lu2022dpm} of the reversed diffusion process:
\begin{equation}
    \mathrm{d}\bm{x}_t = \left[f(t) \bm{x}_t -g^2(t) \nabla_{\bm{x}_t}\log p_t(\bm{x}_t) \right] \mathrm{d}t + g(t) \,\mathrm{d}\widetilde{\bm{w}}_t,
    \label{eq:reverse_sde}
\end{equation}
where $\widetilde{\bm{w}}_t$ is the reverse-time standard Wiener process, and
\begin{equation}
    f(t) = \frac{\mathrm{d}\log\alpha_t}{\mathrm{d}t}, \quad
    g^2(t) = -\sigma_t^2 \frac{\mathrm{d}\log(\alpha_t/\sigma_t)}{\mathrm{d}t}.
    \label{eq:reverse_sde_param}
\end{equation}
A score model $\bm{s}_\theta(\bm{x_t}, t)$ is trained to estimate the score $\nabla_{\bm{x}_t}\log p_t(\bm{x}_t)$ in \cref{eq:reverse_sde} via denoising score matching \cite{vincent2011connection}:
\begin{equation}
    \min_{\theta} \mathbb{E}_{t,\bm{\epsilon},\bm{x}_0,\bm{x}_t} \left[ \| \epsilon + \sigma_t\bm{s}_\theta(\bm{x_t}, t) \|_2^2 \right].
\end{equation}

\vspace{0.5em}
\noindent
\textbf{Degradation Diffusion.}
Previous works \cite{ai2024multimodal, yu2024scaling, lin2023diffbir, jiang2023autodir} model the distribution of HQ images $\bm{x}_\text{HQ}$ with diffusion models guided by LQ images $\bm{x}_\text{LQ}$, so that $\bm{x}_0 = \bm{x}_\text{HQ}$ in \cref{eq:diffusion}. We call them diffusion in image space. We found their formulation inefficient for instruction-guided image restoration, wherein our visual instructions primarily describe the degradations and are loosely correlated with HQ image semantics. Instead, we formulate our visual instruction guided diffusion model in \textit{degradation space}.

Formally, denote $\mathcal{T}_{\bm{v}}(\bm{x}_\text{LQ})$ to be the ``restored'' result of LQ image $\bm{x}_\text{LQ}$ induced by visual instruction $\bm{v}$. That is, $\mathcal{T}_{\bm{v}}(\bm{x}_\text{LQ})$ is the ``cleaner'' version of $\bm{x}_\text{LQ}$ where the degradations are removed \textit{if and only if} they are indicated by $\bm{v}$, and $\mathcal{T}_{\bm{v}}(\bm{x}_\text{LQ}) = \bm{x}_\text{HQ}$ when $\bm{v}$ describes all degradations in $\bm{x}_\text{LQ}$.
The degradation space $\mathcal{D}_{\bm{v}}$ induced by $\bm{v}$ is defined as the set of difference vectors between LQ image set $\mathcal{S}_\text{LQ}$ and their corresponding $\mathcal{T}_{\bm{v}}(\bm{x}_\text{LQ})$: 
\begin{equation}
\mathcal{D}_{\bm{v}}:= \left\{ \bm{x}_\text{LQ} - \mathcal{T}_{\bm{v}}(\bm{x}_\text{LQ}) \,\big\vert\, \bm{x}_\text{LQ} \in \mathcal{S}_\text{LQ} \right\}.
\end{equation}
The degradation diffusion diffuses $\mathcal{D}_{\bm{v}}$ into the support set of standard Gaussian noise $\mathcal{N}(\bm{0}, \bm{I})$. The degradation diffusion is conditioned on both LQ images (HQ images are unavailable during inference) and visual instruction $\bm{v}$:
\begin{equation}
\begin{split}
    & p_t(\bm{y}_t|\bm{y}_0, \bm{x}_\text{LQ}, \bm{v}) = \mathcal{N}(\alpha_t \bm{y}_0, \sigma_t^2 \bm{I}) \\
    & \text{where} \,\, \bm{y}_0 = \bm{x}_\text{LQ} - \mathcal{T}_{\bm{v}}(\bm{x}_\text{LQ}), \,\, 0 \le t \le 1.
\end{split}
\end{equation}
We train the conditioned score model $\bm{s}_\theta(\bm{y_t}, t, \bm{x}_\text{LQ}, \bm{v})$ to estimate the conditioned score $\nabla_{\bm{y}_t}\log p_t(\bm{y}_t | \bm{x}_\text{LQ}, \bm{v})$ via the following objective:
\begin{equation}
    \min_{\theta} \mathbb{E}_{t,\bm{\epsilon},\bm{x}_\text{LQ},\bm{v},\bm{x}_t} \left[ \| \epsilon + \sigma_t \bm{s}_\theta(\bm{y_t}, t, \bm{x}_\text{LQ}, \bm{v}) \|_2^2 \right].
    \label{eq:defusion_objective}
\end{equation}
After training, we numerically solve the reverse time SDE for $t=1\to 0$ from $\bm{y}_1 \sim \mathcal{N}(\bm{0},\bm{I})$ to obtain $\hat{\bm{y}}_0$ as
\begin{equation}
    \mathrm{d}\bm{y}_t = \left[f(t) \bm{y}_t -g^2(t) \bm{s}_\theta(\bm{y_t}, t, \bm{x}_\text{LQ}, \bm{v}) \right] \mathrm{d}t + g(t) \,\mathrm{d}\widetilde{\bm{w}}_t,
\end{equation}
where $f,g$ are defined by \cref{eq:reverse_sde_param}. The final restored image is then computed as $\hat{\mathcal{T}}_{\bm{v}}(\bm{x}_\text{LQ}) = \bm{x}_\text{LQ} - \hat{\bm{y}}_0$.

Our diffusion formulation in degradation space has several advantages over previous diffusion in image space \cite{ai2024multimodal, yu2024scaling, lin2023diffbir, jiang2023autodir}: 1) it achieves better restoration because $\bm{y}_t$ is highly correlated with degradations; 2) training is more stable since the distribution of $\bm{y}_t$ is more consistent than images', which also reduces required model capacity; 3) our formulation \textit{does nothing} when $\bm{v}$ does not match the actual degradation in LQ images, leading to better controllability.
Besides, unlike \cite{liu2024residual, shi2023resfusion, yue2024resshift}, our formulation has no special requirement for $\alpha_t, \sigma_t$; thus, we can adapt any pretrained diffusion model into our formulation with minimal efforts.

\subsection{Conditioning}
\label{subsec:conditioning}

The score model $\bm{s}_\theta(\bm{y_t}, t, \bm{x}_\text{LQ}, \bm{v})$ trained by \cref{eq:defusion_objective} is conditioned on the LQ image $\bm{x}_\text{LQ}$ and visual instruction $\bm{v}$. We implement $\bm{s}_\theta$ with U-Net \cite{ronneberger2015u} and introduce our conditioning mechanisms in this section.

\vspace{0.5em}
\noindent
\textbf{Image Restoration Bridge (IRB).}
The IRB introduces the conditioning on LQ images. IRB first embeds LQ images into hierarchical feature maps of different resolutions with a lightweight convolutional network \cite{mou2024t2i}; each resolution level of feature maps corresponds to that of the diffusion U-Net. The features are injected into the U-Net via AdaLN-Zero \cite{peebles2023scalable}. We found no benefit in concatenating the LQ image with the U-Net input \cite{brooks2023instructpix2pix}. Unlike previous works \cite{yu2024scaling, lin2023diffbir, ai2024multimodal}, the performance deteriorates if we inject the features into the U-Net decoder with ControlNet \cite{zhang2023adding} or a similar adapter. We conjecture the reason is that the prediction in degradation space is dissimilar to the LQ images, leading to diverging features in the U-Net encoder.

\vspace{0.5em}
\noindent
\textbf{Visual Instruction Adapter (VIA).}
Visual instructions control the diffusion U-Net through the VIA. Inspired by IP-Adapter \cite{ye2023ip}, we copy the cross-attention layer in the pretrained text-to-image diffusion U-Net. The query of the copied cross-attention is the (flattened) U-Net feature map, and the key and value are the degradation codes $\bm{z}_v$ encoded from the visual instructions $\bm{v}$ (see \cref{subsec:visual_instructions}). This preserves the text-conditioning ability of the pretrained model, allowing the injection of text descriptions as additional guidance.

\begin{table*}[htp]
\caption{Comparison with state-of-the-art task-specific methods and all-in-one methods on 8 tasks. General IR models trained under the all-in-one setting are marked with a symbol *. The best and second-best performances are in red and bold font, with the top 3 with a light black background.}
\renewcommand\arraystretch{1.30}
\resizebox{\textwidth}{!}{%
\begin{tabular}{cccccccccccc}
\hline
\multicolumn{3}{c|}{\textbf{\begin{tabular}[c]{@{}c@{}}Motion Deblur\\ (Gopro \cite{nah2017deep})\end{tabular}}} &
  \multicolumn{3}{c|}{\textbf{\begin{tabular}[c]{@{}c@{}}Defocus Deblur\\ (DPDD \cite{abuolaim2020defocus})\end{tabular}}} &
  \multicolumn{3}{c|}{\textbf{\begin{tabular}[c]{@{}c@{}}Desnowing\\ (Snow100K-L \cite{liu2018desnownet})\end{tabular}}} &
  \multicolumn{3}{c}{\textbf{\begin{tabular}[c]{@{}c@{}}Dehazing\\ (Dense-Haze \cite{ancuti2019dense})\end{tabular}}} \\ \hline
\multicolumn{1}{c|}{Method} &
  PSNR &
  \multicolumn{1}{c|}{SSIM} &
  \multicolumn{1}{c|}{Method} &
  PSNR &
  \multicolumn{1}{c|}{SSIM} &
  \multicolumn{1}{c|}{Method} &
  PSNR &
  \multicolumn{1}{c|}{SSIM} &
  \multicolumn{1}{c|}{Method} &
  PSNR &
  SSIM \\ \hline
\textbf{Task Specific} &
   &
   &
   &
   &
   &
   &
   &
   &
   &
   &
   \\ \hline
\multicolumn{1}{c|}{MPRNet\cite{zamir2021multi}} &
  32.66 &
  \multicolumn{1}{c|}{0.959} &
  \multicolumn{1}{c|}{DRBNet\cite{ruan2022learning}} &
  25.73 &
  \multicolumn{1}{c|}{0.791} &
  \multicolumn{1}{c|}{DesnowNet\cite{liu2018desnownet}} &
  27.17 &
  \multicolumn{1}{c|}{0.898} &
  \multicolumn{1}{c|}{AECRNet\cite{wu2021contrastive}} &
  15.80 &
  0.466 \\
\multicolumn{1}{c|}{Restormer\cite{zamir2022restormer}} &
  32.92 &
  \multicolumn{1}{c|}{0.961} &
  \multicolumn{1}{c|}{Restormer\cite{zamir2022restormer}} &
  25.98 &
  \multicolumn{1}{c|}{0.811} &
  \multicolumn{1}{c|}{DDMSNet\cite{zhang2021deep}} &
  28.85 &
  \multicolumn{1}{c|}{0.877} &
  \multicolumn{1}{c|}{FocalNet\cite{cui2023focal}} &
  \cellcolor[HTML]{EFEFEF}{\ul \textbf{17.07}} &
  \cellcolor[HTML]{EFEFEF}{\ul \textbf{0.630}} \\
\multicolumn{1}{c|}{Stripformer\cite{tsai2022stripformer}} &
  \cellcolor[HTML]{EFEFEF}33.08 &
  \multicolumn{1}{c|}{\cellcolor[HTML]{EFEFEF}0.962} &
  \multicolumn{1}{c|}{NRKNet\cite{quan2023neumann}} &
  26.11 &
  \multicolumn{1}{c|}{0.810} &
  \multicolumn{1}{c|}{DRT\cite{liang2022drt}} &
  29.56 &
  \multicolumn{1}{c|}{0.892} &
  \multicolumn{1}{c|}{DeHamer\cite{guo2022image}} &
  16.62 &
  0.560 \\
\multicolumn{1}{c|}{DiffIR\cite{xia2023diffir}} &
  \cellcolor[HTML]{EFEFEF}{\ul \textbf{33.20}} &
  \multicolumn{1}{c|}{\cellcolor[HTML]{EFEFEF}{\ul \textbf{0.963}}} &
  \multicolumn{1}{c|}{FocalNet\cite{cui2023focal}} &
  26.18 &
  \multicolumn{1}{c|}{0.808} &
  \multicolumn{1}{c|}{WeatherDiff\cite{ozdenizci2023restoring}} &
  30.43 &
  \multicolumn{1}{c|}{\cellcolor[HTML]{EFEFEF}0.915} &
  \multicolumn{1}{c|}{MB-Taylor\cite{qiu2023mb}} &
  16.44 &
  0.566 \\ \hline
\textbf{All in One} &
   &
   &
   &
   &
   &
   &
   &
   &
   &
   &
   \\ \hline
\multicolumn{1}{c|}{AirNet\cite{li2022all}} &
  28.31 &
  \multicolumn{1}{c|}{0.910} &
  \multicolumn{1}{c|}{AirNet\cite{li2022all}} &
  25.37 &
  \multicolumn{1}{c|}{0.770} &
  \multicolumn{1}{c|}{AirNet\cite{li2022all}} &
  30.14 &
  \multicolumn{1}{c|}{0.907} &
  \multicolumn{1}{c|}{AirNet\cite{li2022all}} &
  15.32 &
  0.562 \\
\multicolumn{1}{c|}{PromptIR\cite{potlapalli2024promptir}} &
  31.02 &
  \multicolumn{1}{c|}{0.938} &
  \multicolumn{1}{c|}{PromptIR\cite{potlapalli2024promptir}} &
  25.66 &
  \multicolumn{1}{c|}{0.791} &
  \multicolumn{1}{c|}{PromptIR\cite{potlapalli2024promptir}} &
  \cellcolor[HTML]{EFEFEF}30.91 &
  \multicolumn{1}{c|}{0.913} &
  \multicolumn{1}{c|}{PromptIR\cite{potlapalli2024promptir}} &
  16.71 &
  \cellcolor[HTML]{EFEFEF}0.580 \\
\multicolumn{1}{c|}{DA-CLIP\cite{luo2023controlling}} &
  27.12 &
  \multicolumn{1}{c|}{0.823} &
  \multicolumn{1}{c|}{DA-CLIP\cite{luo2023controlling}} &
  24.91 &
  \multicolumn{1}{c|}{0.749} &
  \multicolumn{1}{c|}{DA-CLIP\cite{luo2023controlling}} &
  28.31 &
  \multicolumn{1}{c|}{0.862} &
  \multicolumn{1}{c|}{TransWeather\cite{valanarasu2022transweather}} &
  15.58 &
  0.569 \\
\multicolumn{1}{c|}{Restormer*\cite{zamir2022restormer}} &
  31.17 &
  \multicolumn{1}{c|}{0.931} &
  \multicolumn{1}{c|}{Restormer*\cite{zamir2022restormer}} &
  \cellcolor[HTML]{EFEFEF}28.08 &
  \multicolumn{1}{c|}{\cellcolor[HTML]{EFEFEF}{\ul \textbf{0.898}}} &
  \multicolumn{1}{c|}{Restormer*\cite{zamir2022restormer}} &
  29.18 &
  \multicolumn{1}{c|}{0.899} &
  \multicolumn{1}{c|}{Restormer*\cite{zamir2022restormer}} &
  16.18 &
  0.560 \\
\multicolumn{1}{c|}{NAFNet*\cite{chen2022simple}} &
  31.44 &
  \multicolumn{1}{c|}{0.939} &
  \multicolumn{1}{c|}{NAFNet*\cite{chen2022simple}} &
  \cellcolor[HTML]{EFEFEF}{\ul \textbf{28.21}} &
  \multicolumn{1}{c|}{\cellcolor[HTML]{EFEFEF}0.892} &
  \multicolumn{1}{c|}{NAFNet*\cite{chen2022simple}} &
  29.21 &
  \multicolumn{1}{c|}{0.896} &
  \multicolumn{1}{c|}{NAFNet*\cite{chen2022simple}} &
  \cellcolor[HTML]{EFEFEF}16.72 &
  0.602 \\
\multicolumn{1}{c|}{MPerceiver\cite{ai2024multimodal}} &
  32.49 &
  \multicolumn{1}{c|}{0.959} &
  \multicolumn{1}{c|}{MPerceiver\cite{ai2024multimodal}} &
  25.88 &
  \multicolumn{1}{c|}{0.803} &
  \multicolumn{1}{c|}{MPerceiver\cite{ai2024multimodal}} &
  \cellcolor[HTML]{EFEFEF}{\ul \textbf{31.02}} &
  \multicolumn{1}{c|}{\cellcolor[HTML]{EFEFEF}{\ul \textbf{0.916}}} &
  \multicolumn{1}{c|}{WGWS-Net\cite{zhang2021plug}} &
  15.50 &
  0.535 \\
\multicolumn{1}{c|}{\textbf{Defusion(Ours)}} &
  \cellcolor[HTML]{EFEFEF}{\color[HTML]{FE0000} \textbf{34.53}} &
  \multicolumn{1}{c|}{\cellcolor[HTML]{EFEFEF}{\color[HTML]{FE0000} \textbf{0.966}}} &
  \multicolumn{1}{c|}{\textbf{Defusion(Ours)}} &
  \cellcolor[HTML]{EFEFEF}{\color[HTML]{FE0000} \textbf{29.68}} &
  \multicolumn{1}{c|}{\cellcolor[HTML]{EFEFEF}{\color[HTML]{FE0000} \textbf{0.922}}} &
  \multicolumn{1}{c|}{\textbf{Defusion(Ours)}} &
  \cellcolor[HTML]{EFEFEF}{\color[HTML]{FE0000} \textbf{32.11}} &
  \multicolumn{1}{c|}{\cellcolor[HTML]{EFEFEF}{\color[HTML]{FE0000} \textbf{0.926}}} &
  \multicolumn{1}{c|}{\textbf{Defusion(Ours)}} &
  \cellcolor[HTML]{EFEFEF}{\color[HTML]{FE0000} \textbf{17.55}} &
  \cellcolor[HTML]{EFEFEF}{\color[HTML]{FE0000} \textbf{0.677}} \\ \hline
\multicolumn{3}{c|}{\textbf{\begin{tabular}[c]{@{}c@{}}Raindrop Removal\\ (RainDrop \cite{qian2018attentive})\end{tabular}}} &
  \multicolumn{3}{c|}{\textbf{\begin{tabular}[c]{@{}c@{}}Deraining\\ (Rain1400 \cite{fu2017removing})\end{tabular}}} &
  \multicolumn{3}{c|}{\textbf{\begin{tabular}[c]{@{}c@{}}Real Denoising\\ (SIDD \cite{abdelhamed2018high})\end{tabular}}} &
  \multicolumn{3}{c}{\textbf{\begin{tabular}[c]{@{}c@{}}Artifact removal\\ (LIVE1 \cite{sheikh2005live})\end{tabular}}} \\ \hline
\multicolumn{1}{c|}{Method} &
  PSNR &
  \multicolumn{1}{c|}{SSIM} &
  \multicolumn{1}{c|}{Method} &
  PSNR &
  \multicolumn{1}{c|}{SSIM} &
  \multicolumn{1}{c|}{Method} &
  PSNR &
  \multicolumn{1}{c|}{SSIM} &
  \multicolumn{1}{c|}{Method} &
  PSNR &
  SSIM \\ \hline
\textbf{Task Specific} &
   &
   &
   &
   &
   &
   &
   &
   &
   &
   &
   \\ \hline
\multicolumn{1}{c|}{AttentGAN\cite{qian2018attentive}} &
  31.59 &
  \multicolumn{1}{c|}{0.917} &
  \multicolumn{1}{c|}{Uformer\cite{wang2022uformer}} &
  32.84 &
  \multicolumn{1}{c|}{0.931} &
  \multicolumn{1}{c|}{MPRNet\cite{zamir2021multi}} &
  39.71 &
  \multicolumn{1}{c|}{0.958} &
  \multicolumn{1}{c|}{QGAC\cite{ehrlich2020quantization}} &
  27.62 &
  0.804 \\
\multicolumn{1}{c|}{Quanetal.\cite{quan2019deep}} &
  31.37 &
  \multicolumn{1}{c|}{0.918} &
  \multicolumn{1}{c|}{Restormer\cite{zamir2022restormer}} &
  \cellcolor[HTML]{EFEFEF}{\ul \textbf{33.68}} &
  \multicolumn{1}{c|}{0.939} &
  \multicolumn{1}{c|}{Uformer\cite{wang2022uformer}} &
  39.89 &
  \multicolumn{1}{c|}{0.960} &
  \multicolumn{1}{c|}{FBCNN\cite{cho2021rethinking}} &
  27.77 &
  0.803 \\
\multicolumn{1}{c|}{IDT\cite{xiao2022image}} &
  31.87 &
  \multicolumn{1}{c|}{0.931} &
  \multicolumn{1}{c|}{DRSformer\cite{chen2023learning}} &
  \cellcolor[HTML]{EFEFEF}33.66 &
  \multicolumn{1}{c|}{0.939} &
  \multicolumn{1}{c|}{Restormer\cite{zamir2022restormer}} &
  40.02 &
  \multicolumn{1}{c|}{0.960} &
  \multicolumn{1}{c|}{SwinIR\cite{yang2019scale}} &
  27.45 &
  0.796 \\
\multicolumn{1}{c|}{UDR-S\cite{chen2023sparse}} &
  \cellcolor[HTML]{EFEFEF}32.64 &
  \multicolumn{1}{c|}{0.942} &
  \multicolumn{1}{c|}{UDR-S2\cite{chen2023sparse}} &
  33.08 &
  \multicolumn{1}{c|}{0.930} &
  \multicolumn{1}{c|}{ART\cite{zhang2022accurate}} &
  39.99 &
  \multicolumn{1}{c|}{0.960} &
  \multicolumn{1}{c|}{DAGN\cite{ma2024sensitivity}} &
  \cellcolor[HTML]{EFEFEF}{\color[HTML]{FE0000} \textbf{27.95}} &
  0.807 \\ \hline
\textbf{All in One} &
   &
   &
   &
   &
   &
   &
   &
   &
   &
   &
   \\ \hline
\multicolumn{1}{c|}{AirNet\cite{li2022all}} &
  31.32 &
  \multicolumn{1}{c|}{0.925} &
  \multicolumn{1}{c|}{AirNet\cite{li2022all}} &
  32.36 &
  \multicolumn{1}{c|}{0.928} &
  \multicolumn{1}{c|}{AirNet\cite{li2022all}} &
  38.32 &
  \multicolumn{1}{c|}{0.945} &
  \multicolumn{1}{c|}{AirNet\cite{li2022all}} &
  15.32 &
  0.562 \\
\multicolumn{1}{c|}{PromptIR\cite{potlapalli2024promptir}} &
  32.03 &
  \multicolumn{1}{c|}{0.938} &
  \multicolumn{1}{c|}{PromptIR\cite{potlapalli2024promptir}} &
  33.26 &
  \multicolumn{1}{c|}{0.935} &
  \multicolumn{1}{c|}{PromptIR\cite{potlapalli2024promptir}} &
  39.52 &
  \multicolumn{1}{c|}{0.954} &
  \multicolumn{1}{c|}{TransWeather\cite{valanarasu2022transweather}} &
  15.58 &
  0.569 \\
\multicolumn{1}{c|}{DA-CLIP\cite{luo2023controlling}} &
  30.44 &
  \multicolumn{1}{c|}{0.880} &
  \multicolumn{1}{c|}{DA-CLIP\cite{luo2023controlling}} &
  29.67 &
  \multicolumn{1}{c|}{0.851} &
  \multicolumn{1}{c|}{DA-CLIP\cite{luo2023controlling}} &
  34.04 &
  \multicolumn{1}{c|}{0.824} &
  \multicolumn{1}{c|}{WGWS-Net\cite{zhang2021plug}} &
  15.50 &
  0.535 \\
\multicolumn{1}{c|}{Restormer*\cite{zamir2022restormer}} &
  31.67 &
  \multicolumn{1}{c|}{\cellcolor[HTML]{EFEFEF}{\ul \textbf{0.958}}} &
  \multicolumn{1}{c|}{Restormer*\cite{zamir2022restormer}} &
  32.61 &
  \multicolumn{1}{c|}{\cellcolor[HTML]{EFEFEF}{\ul \textbf{0.944}}} &
  \multicolumn{1}{c|}{Restormer*\cite{zamir2022restormer}} &
  \cellcolor[HTML]{EFEFEF}{\ul \textbf{40.60}} &
  \multicolumn{1}{c|}{0.955} &
  \multicolumn{1}{c|}{Restormer*\cite{zamir2022restormer}} &
  26.77 &
  \cellcolor[HTML]{EFEFEF}{\ul \textbf{0.827}} \\
\multicolumn{1}{c|}{NAFNet*\cite{chen2022simple}} &
  31.03 &
  \multicolumn{1}{c|}{\cellcolor[HTML]{EFEFEF}0.952} &
  \multicolumn{1}{c|}{NAFNet*\cite{chen2022simple}} &
  32.38 &
  \multicolumn{1}{c|}{\cellcolor[HTML]{EFEFEF}0.943} &
  \multicolumn{1}{c|}{NAFNet*\cite{chen2022simple}} &
  \cellcolor[HTML]{EFEFEF}40.22 &
  \multicolumn{1}{c|}{0.952} &
  \multicolumn{1}{c|}{NAFNet*\cite{chen2022simple}} &
  26.79 &
  \cellcolor[HTML]{EFEFEF}{\ul \textbf{0.827}} \\
\multicolumn{1}{c|}{MPerceiver\cite{ai2024multimodal}} &
  \cellcolor[HTML]{EFEFEF}{\ul \textbf{33.21}} &
  \multicolumn{1}{c|}{0.929} &
  \multicolumn{1}{c|}{MPerceiver\cite{ai2024multimodal}} &
  33.40 &
  \multicolumn{1}{c|}{0.937} &
  \multicolumn{1}{c|}{MPerceiver\cite{ai2024multimodal}} &
  39.96 &
  \multicolumn{1}{c|}{0.959} &
  \multicolumn{1}{c|}{MPerceiver\cite{ai2024multimodal}} &
  \cellcolor[HTML]{EFEFEF}{\ul \textbf{27.79}} &
  0.804 \\
\multicolumn{1}{c|}{\textbf{Defusion(Ours)}} &
  \cellcolor[HTML]{EFEFEF}{\color[HTML]{FE0000} \textbf{33.81}} &
  \multicolumn{1}{c|}{\cellcolor[HTML]{EFEFEF}{\color[HTML]{FE0000} \textbf{0.967}}} &
  \multicolumn{1}{c|}{\textbf{Defusion(Ours)}} &
  \cellcolor[HTML]{EFEFEF}{\color[HTML]{FE0000} \textbf{34.09}} &
  \multicolumn{1}{c|}{\cellcolor[HTML]{EFEFEF}{\color[HTML]{FE0000} \textbf{0.954}}} &
  \multicolumn{1}{c|}{\textbf{Defusion(Ours)}} &
  \cellcolor[HTML]{EFEFEF}{\color[HTML]{FE0000} \textbf{40.89}} &
  \multicolumn{1}{c|}{0.954} &
  \multicolumn{1}{c|}{\textbf{Defusion(Ours)}} &
  \cellcolor[HTML]{EFEFEF}27.77 &
  \cellcolor[HTML]{EFEFEF}{\color[HTML]{FE0000} \textbf{0.832}} \\ \hline
\end{tabular}%
}
\label{table:aio}
\vspace{-0.4cm}
\end{table*}

\begin{figure*}[ht]
    \centering
    \includegraphics[width=\textwidth]{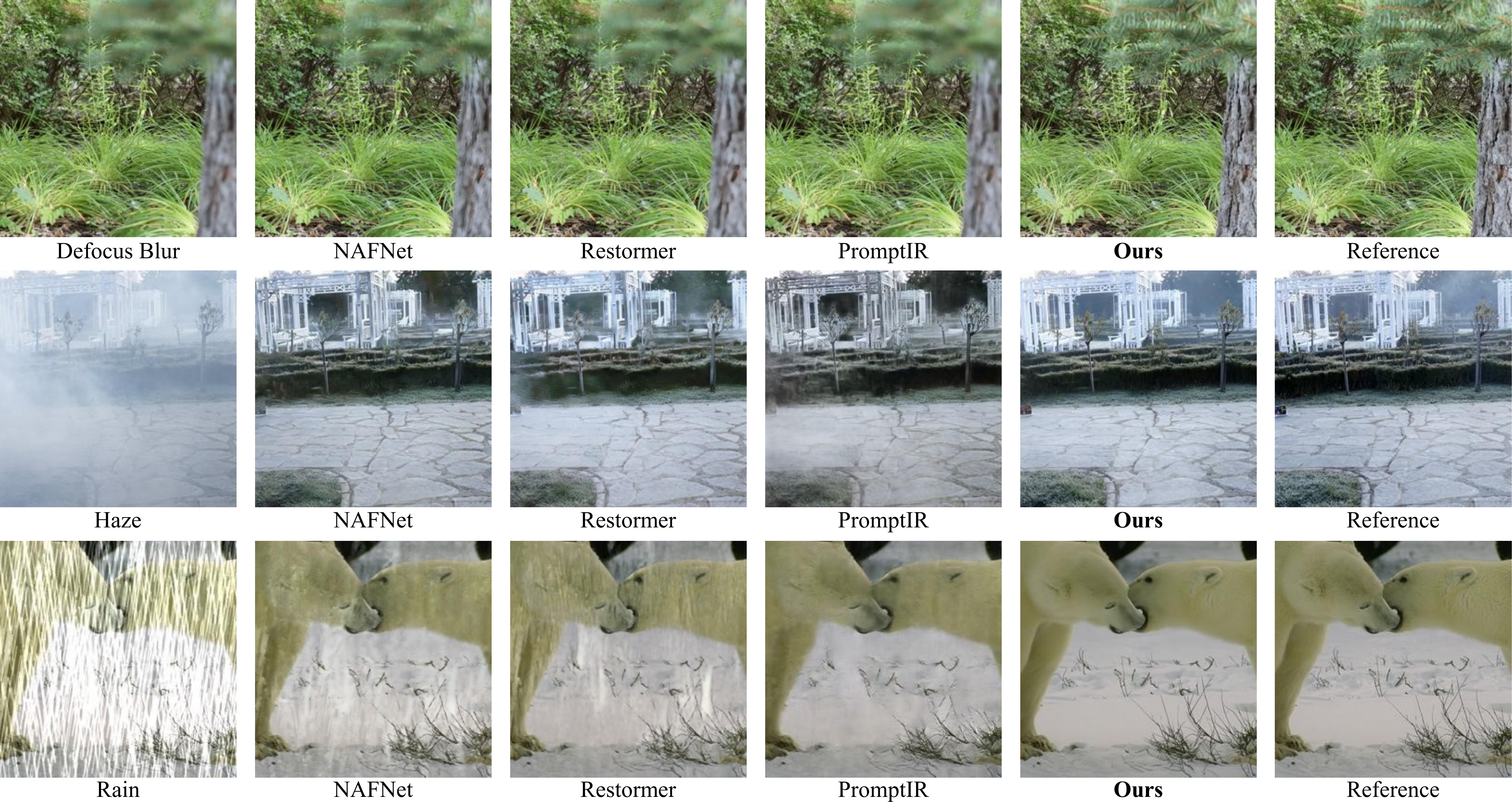}
    \caption{Visual results on dehazing, draining, and deblurring on synthetic datasets .}
    \label{fig:aio}
    \vspace{-0.2cm}
\end{figure*}

\begin{table*}[]
\caption{Real-world datasets results. General IR models trained under the all-in-one training set are marked with a symbol *. The best and second-best performances are in red and bold font, with the top 3 with a light black background.}
\renewcommand\arraystretch{1.30}
\resizebox{\textwidth}{!}{%
\begin{tabular}{cccccccccccc}
\hline
\multicolumn{3}{c|}{\textbf{Dehazing (NH-HAZE \cite{ancuti2020nh})}} &
  \multicolumn{3}{c|}{\textbf{Deraining (LHP \cite{guo2023sky})}} &
  \multicolumn{3}{c|}{\textbf{Desnowing (RealSnow \cite{zhu2023learning})}} &
  \multicolumn{3}{c}{\textbf{Motion Deblur (RealBlur-J \cite{rim2020real})}} \\ \hline
\multicolumn{1}{c|}{Method} &
  PSNR &
  \multicolumn{1}{c|}{SSIM} &
  \multicolumn{1}{c|}{Method} &
  PSNR &
  \multicolumn{1}{c|}{SSIM} &
  \multicolumn{1}{c|}{Method} &
  PSNR &
  \multicolumn{1}{c|}{SSIM} &
  \multicolumn{1}{c|}{Method} &
  PSNR &
  SSIM \\ \hline
\textbf{Task Specific} &
   &
   &
   &
   &
   &
   &
   &
   &
   &
   &
   \\ \hline
\multicolumn{1}{c|}{AECRNet\cite{wu2021contrastive}} &
  19.88 &
  \multicolumn{1}{c|}{0.625} &
  \multicolumn{1}{c|}{MUSS\cite{huang2022memory}} &
  30.02 &
  \multicolumn{1}{c|}{0.886} &
  \multicolumn{1}{c|}{MIRNetv2\cite{zamir2022learning}} &
  \cellcolor[HTML]{EFEFEF}31.39 &
  \multicolumn{1}{c|}{0.916} &
  \multicolumn{1}{c|}{MPRNet\cite{zamir2021multi}} &
  28.70 &
  0.873 \\
\multicolumn{1}{c|}{SGID\cite{bai2022self}} &
  16.23 &
  \multicolumn{1}{c|}{0.606} &
  \multicolumn{1}{c|}{Restormer\cite{zamir2022restormer}} &
  29.72 &
  \multicolumn{1}{c|}{0.889} &
  \multicolumn{1}{c|}{ART\cite{zhang2022accurate}} &
  31.05 &
  \multicolumn{1}{c|}{0.913} &
  \multicolumn{1}{c|}{Restormer\cite{zamir2022restormer}} &
  28.96 &
  0.879 \\
\multicolumn{1}{c|}{DeHamer\cite{guo2022image}} &
  20.66 &
  \multicolumn{1}{c|}{0.684} &
  \multicolumn{1}{c|}{DRSformer\cite{chen2023learning}} &
  30.04 &
  \multicolumn{1}{c|}{0.895} &
  \multicolumn{1}{c|}{Restormer\cite{zamir2022restormer}} &
  31.38 &
  \multicolumn{1}{c|}{0.923} &
  \multicolumn{1}{c|}{Stripformer\cite{tsai2022stripformer}} &
  28.82 &
  0.876 \\
\multicolumn{1}{c|}{MB-Taylor\cite{qiu2023mb}} &
  19.43 &
  \multicolumn{1}{c|}{0.638} &
  \multicolumn{1}{c|}{UDR-S2\cite{chen2023sparse}} &
  28.59 &
  \multicolumn{1}{c|}{0.884} &
  \multicolumn{1}{c|}{NAFNet\cite{chen2022simple}} &
  31.44 &
  \multicolumn{1}{c|}{0.919} &
  \multicolumn{1}{c|}{DiffIR\cite{xia2023diffir}} &
  29.06 &
  0.882 \\ \hline
\textbf{All in One} &
   &
   &
   &
   &
   &
   &
   &
   &
   &
   &
   \\ \hline
\multicolumn{1}{c|}{AirNet\cite{li2022all}} &
  17.63 &
  \multicolumn{1}{c|}{0.615} &
  \multicolumn{1}{c|}{AirNet\cite{li2022all}} &
  31.73 &
  \multicolumn{1}{c|}{0.889} &
  \multicolumn{1}{c|}{AirNet\cite{li2022all}} &
  31.02 &
  \multicolumn{1}{c|}{0.923} &
  \multicolumn{1}{c|}{AirNet\cite{li2022all}} &
  27.91 &
  0.834 \\
\multicolumn{1}{c|}{TransWeather\cite{valanarasu2022transweather}} &
  19.19 &
  \multicolumn{1}{c|}{0.644} &
  \multicolumn{1}{c|}{TransWeather\cite{valanarasu2022transweather}} &
  29.87 &
  \multicolumn{1}{c|}{0.867} &
  \multicolumn{1}{c|}{TransWeather\cite{valanarasu2022transweather}} &
  31.13 &
  \multicolumn{1}{c|}{\cellcolor[HTML]{EFEFEF}0.922} &
  \multicolumn{1}{c|}{TransWeather\cite{valanarasu2022transweather}} &
  28.03 &
  0.837 \\
\multicolumn{1}{c|}{WGWS-Net\cite{zhang2021plug}} &
  18.39 &
  \multicolumn{1}{c|}{0.639} &
  \multicolumn{1}{c|}{WGWS-Net\cite{zhang2021plug}} &
  \cellcolor[HTML]{EFEFEF}30.77 &
  \multicolumn{1}{c|}{0.885} &
  \multicolumn{1}{c|}{WGWS-Net\cite{zhang2021plug}} &
  31.37 &
  \multicolumn{1}{c|}{0.919} &
  \multicolumn{1}{c|}{WGWS-Net\cite{zhang2021plug}} &
  28.10 &
  0.838 \\
\multicolumn{1}{c|}{Restormer*\cite{zamir2022restormer}} &
  \cellcolor[HTML]{EFEFEF}{\ul \textbf{22.57}} &
  \multicolumn{1}{c|}{\cellcolor[HTML]{EFEFEF}{\ul \textbf{0.795}}} &
  \multicolumn{1}{c|}{Restormer*\cite{zamir2022restormer}} &
  29.87 &
  \multicolumn{1}{c|}{\cellcolor[HTML]{EFEFEF}{\ul \textbf{0.928}}} &
  \multicolumn{1}{c|}{Restormer*\cite{zamir2022restormer}} &
  29.47 &
  \multicolumn{1}{c|}{0.889} &
  \multicolumn{1}{c|}{Restormer*\cite{zamir2022restormer}} &
  \cellcolor[HTML]{EFEFEF}{\ul \textbf{30.91}} &
  \cellcolor[HTML]{EFEFEF}{\ul \textbf{0.930}} \\
\multicolumn{1}{c|}{NAFNet*\cite{chen2022simple}} &
  21.66 &
  \multicolumn{1}{c|}{\cellcolor[HTML]{EFEFEF}0.783} &
  \multicolumn{1}{c|}{NAFNet*\cite{chen2022simple}} &
  27.95 &
  \multicolumn{1}{c|}{\cellcolor[HTML]{EFEFEF}0.917} &
  \multicolumn{1}{c|}{NAFNet*\cite{chen2022simple}} &
  28.15 &
  \multicolumn{1}{c|}{0.858} &
  \multicolumn{1}{c|}{NAFNet*\cite{chen2022simple}} &
  \cellcolor[HTML]{EFEFEF}30.80 &
  \cellcolor[HTML]{EFEFEF}0.928 \\
\multicolumn{1}{c|}{MPerceiver\cite{ai2024multimodal}} &
  \cellcolor[HTML]{EFEFEF}21.97 &
  \multicolumn{1}{c|}{0.646} &
  \multicolumn{1}{c|}{MPerceiver\cite{ai2024multimodal}} &
  \cellcolor[HTML]{EFEFEF}{\ul \textbf{32.07}} &
  \multicolumn{1}{c|}{0.889} &
  \multicolumn{1}{c|}{MPerceiver\cite{ai2024multimodal}} &
  \cellcolor[HTML]{EFEFEF}{\ul \textbf{31.45}} &
  \multicolumn{1}{c|}{\cellcolor[HTML]{EFEFEF}{\ul \textbf{0.924}}} &
  \multicolumn{1}{c|}{MPerceiver\cite{ai2024multimodal}} &
  29.13 &
  0.881 \\
\multicolumn{1}{c|}{\textbf{Defusion(Ours)}} &
  \cellcolor[HTML]{EFEFEF}{\color[HTML]{FE0000} \textbf{22.87}} &
  \multicolumn{1}{c|}{\cellcolor[HTML]{EFEFEF}{\color[HTML]{FE0000} \textbf{0.853}}} &
  \multicolumn{1}{c|}{\textbf{Defusion(Ours)}} &
  \cellcolor[HTML]{EFEFEF}{\color[HTML]{FE0000} \textbf{32.98}} &
  \multicolumn{1}{c|}{\cellcolor[HTML]{EFEFEF}{\color[HTML]{FE0000} \textbf{0.947}}} &
  \multicolumn{1}{c|}{\textbf{Defusion(Ours)}} &
  \cellcolor[HTML]{EFEFEF}{\color[HTML]{FE0000} \textbf{32.73}} &
  \multicolumn{1}{c|}{\cellcolor[HTML]{EFEFEF}{\color[HTML]{FE0000} \textbf{0.930}}} &
  \multicolumn{1}{c|}{\textbf{Defusion(Ours)}} &
  \cellcolor[HTML]{EFEFEF}{\color[HTML]{FE0000} \textbf{32.39}} &
  \cellcolor[HTML]{EFEFEF}{\color[HTML]{FE0000} \textbf{0.945}} \\ \hline
\end{tabular}%
}
\vspace{-0.2cm}
\end{table*}

\begin{figure*}[ht]
    \centering
    \includegraphics[width=\textwidth]{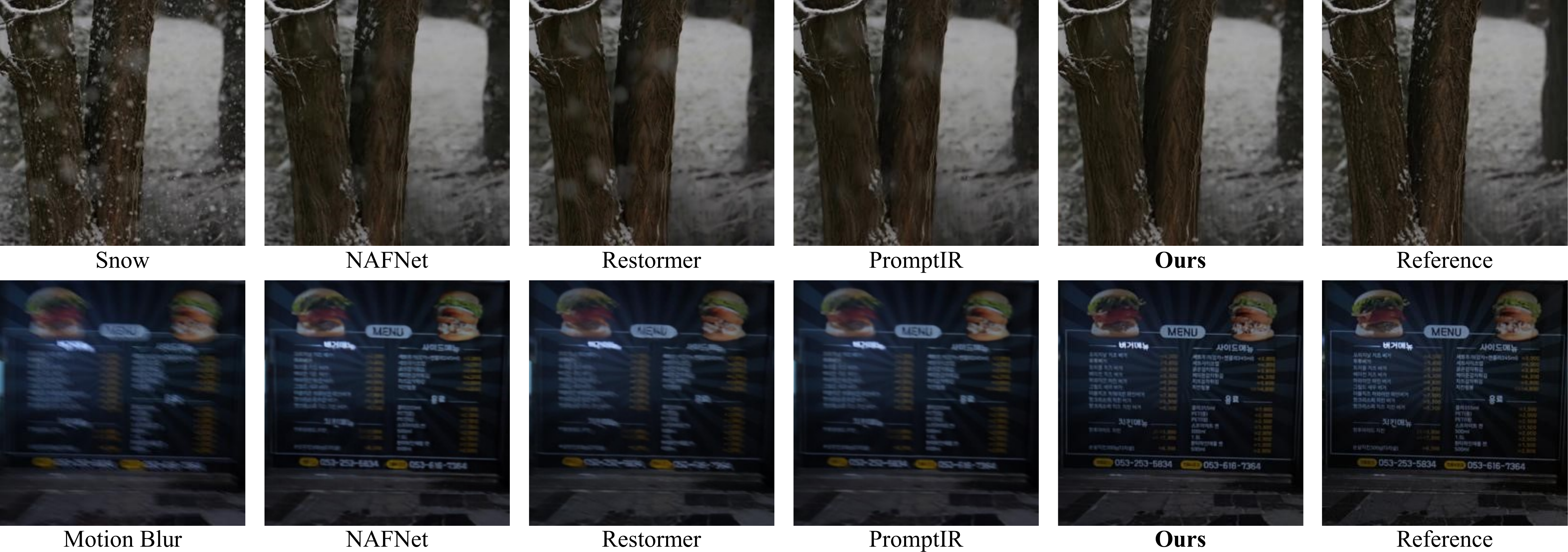}
    \caption{Real-world visual results on desnowing and deblurring.}
    \label{fig:real}
    \vspace{-0.2cm}
\end{figure*}

\begin{table*}[htp]
\caption{Synthesized mixed distortion dataset. The $\rightarrow$ indicates the order in which distortions are added. General IR models trained under the all-in-one training set are marked with a symbol *.}
\vspace{-0.2cm}
\renewcommand\arraystretch{1.30}
\resizebox{\textwidth}{!}{%
\begin{tabular}{c|cc|cc|cc|cc|cc|cc}
\hline
\multirow{2}{*}{\textbf{Method}} & \multicolumn{2}{c|}{\textbf{Rain$\rightarrow$\rm{Noise}$\rightarrow$\rm{Snow}}} & \multicolumn{2}{c|}{\textbf{Rain$\rightarrow$\rm{Snow}$\rightarrow$ \rm{Noise}}} & \multicolumn{2}{c|}{\textbf{Noise$\rightarrow$\rm{Snow}$\rightarrow$ \rm{Rain}}} & \multicolumn{2}{c|}{\textbf{Noise$\rightarrow$\rm{Rain}$\rightarrow$ \rm{Snow}}} & \multicolumn{2}{c|}{\textbf{Snow$\rightarrow$\rm{Rain}$\rightarrow$ \rm{Noise}}} & \multicolumn{2}{c}{\textbf{Snow$\rightarrow$\rm{Noise}$\rightarrow$ \rm{Rain}}} \\ \cline{2-13} 
 & PSNR & SSIM & PSNR & SSIM & PSNR & SSIM & PSNR & SSIM & PSNR & SSIM & PSNR & SSIM \\ \hline
WeatherDiff\cite{ozdenizci2023restoring} & 14.744 & 0.363 & 14.693 & 0.361 & 14.913 & 0.397 & 14.839 & 0.393 & 14.688 & 0.359 & 14.843 & 0.395 \\
PromptIR\cite{potlapalli2024promptir} & 13.976 & 0.471 & 14.212 & 0.492 & 14.544 & 0.504 & 14.357 & 0.494 & 14.553 & 0.510 & 14.615 & 0.513 \\
Restormer*\cite{zamir2022restormer} & 24.114 & 0.766 & 23.895 & 0.760 & 25.168 & 0.802 & 25.022 & 0.798 & 24.144 & 0.766 & 24.814 & 0.793 \\
NAFNet*\cite{chen2022simple} & 23.815 & 0.755 & 23.775 & 0.755 & 24.203 & 0.776 & 24.204 & 0.773 & 23.564 & 0.745 & 24.106 & 0.772 \\ \hline
\textbf{Defusion(Ours)} & \textbf{25.511} & \textbf{0.812} & \textbf{25.228} & \textbf{0.807} & \textbf{27.008} & \textbf{0.855} & \textbf{26.600} & \textbf{0.847} & \textbf{25.052} & \textbf{0.802} & \textbf{26.079} & \textbf{0.837} \\ \hline
\end{tabular}%
}
\label{table:mix}
\end{table*}

\begin{table}[]
\caption{UnderWater datasets results. Method trained under the all-in-one training set are marked with a symbol *.}
\vspace{-0.2cm}
\renewcommand\arraystretch{1.30}
\resizebox{\columnwidth}{!}{%
\begin{tabular}{c|ccc|ccc}
\hline
\multirow{2}{*}{\textbf{Method}} & \multicolumn{3}{c|}{\textbf{EUVP} \cite{islam2020fast}} & \multicolumn{3}{c}{\textbf{TURBID \cite{duarte2016dataset}} } \\ \cline{2-7} 
 & PSNR$\uparrow$ & SSIM$\uparrow$ & LPIPS$\downarrow$ & PSNR$\uparrow$ & SSIM$\uparrow$ & LPIPS$\downarrow$ \\ \hline
PromptIR\cite{potlapalli2024promptir} & 20.046 & 0.799 & 0.296 & 15.324 & 0.593 & 0.324 \\
Restormer*\cite{zamir2022restormer} & 19.764 & 0.788 & 0.316 & 17.420 & 0.648 & 0.282 \\
NAFNet*\cite{chen2022simple} & 19.661 & 0.779 & 0.314 & 17.571 & 0.672 & 0.291 \\ \hline
\textbf{Defusion(Ours)} & \textbf{21.233} & \textbf{0.814} & \textbf{0.269} & \textbf{18.261} & \textbf{0.659} & \textbf{0.283} \\ \hline
\end{tabular}%
}
\label{table:underwater}
\vspace{-0.2cm}
\end{table}

\begin{table}[]
\caption{Ablation study. The metrics are reported on the average of deraining, dehazing and desnowing.}
\vspace{-0.2cm}
\renewcommand\arraystretch{1.30}
\resizebox{\columnwidth}{!}{%
\begin{tabular}{c|ccc}
\hline
\multirow{2}{*}{\textbf{Method}} & \multicolumn{3}{c}{\textbf{Average}} \\ \cline{2-4} 
 & PSNR$\uparrow$ & SSIM$\uparrow$ & LPIPS$\downarrow$ \\ \hline
Defusion(Ours) & 33.83 & 0.951 & 0.025 \\
- replace with Naive Diffusion & 31.72 & 0.918 & 0.086 \\ \hline
- replace with Blank Grounding & 30.89 & 0.903 & 0.102 \\
- replace with Simple Grounding & 31.64 & 0.927 & 0.044 \\ \hline
- replace with Text Instruction & 32.57 & 0.929 & 0.041 \\
- without Instruction & 30.86 & 0.903 & 0.101 \\ \hline
\end{tabular}%
}
\label{table:abl}
\vspace{-0.4cm}
\end{table}

\section{Experiments}
\label{sec:exp}

\subsection{Experiment Settings}

\noindent
\textbf{Datasets and Metrics.}
We combined various image restoration datasets to form the training set, following \cite{zhang2023ingredient}. This includes datasets for deraining such as Rain1400 \cite{fu2017removing}, Outdoor-Rain \cite{li2019heavy}, and LHP \cite{guo2023sky}; dehazing datasets like RESIDE \cite{li2018benchmarking}, and Dense-Haze \cite{ancuti2019dense}; desnowing datasets such as Snow100K \cite{liu2018desnownet} and RealSnow \cite{zhu2023learning}; raindrop removal datasets like RainDrop \cite{qian2018attentive} and RainDS \cite{quan2021removing}; real denoising with SIDD \cite{abdelhamed2018high}; motion deblurring with GoPro \cite{nah2017deep} and RealBlur \cite{rim2020real}; defocus deblurring with DPDD \cite{abuolaim2020defocus}.

During the evaluation, we used Dense-Haze \cite{ancuti2019dense} for dehazing, Rain1400 \cite{fu2017removing} for deraining, GoPro \cite{nah2017deep} for motion deblurring, RainDrop \cite{qian2018attentive} for raindrop removal, DPDD \cite{abuolaim2020defocus} for defocus deblurring, Snow100K \cite{liu2018desnownet} for desnowing, SIDD \cite{abdelhamed2018high} for real denoising, and LIVE1 \cite{sheikh2005live} for JPEG artifact removal (with quality factor $q=10$) on the synthetic datasets.
For real-world degradation datasets, we evaluated our model on the NH-HAZE \cite{ancuti2020nh} dataset, LHP \cite{guo2023sky} dataset, RealBlur \cite{rim2020real} dataset, and the RealSnow \cite{zhu2023learning} dataset.

Moreover, following the method in \cite{park2023all}, we constructed a mixed distortion dataset on the WED \cite{ma2016waterloo} dataset using rain, snow, and noise distortions. We used these diverse relationships in this dataset to evaluate the model’s ability to handle complex distortions. We also introduced underwater datasets EUVP \cite{islam2020fast} and TURBID \cite{duarte2016dataset} to examine the ability of our method to deal with composite distortions in special environments.
We used PSNR and SSIM \cite{wang2004image} as distortion metrics, LPIPS \cite{zhang2018unreasonable} as perceptual metrics.

\noindent
\textbf{Implementation Details.}
Defusion was trained using the Adam optimizer \cite{kingma2014adam} with learning rates of $\beta_1 = 0.9$ and $\beta_2 = 0.999$, starting with an initial learning rate of 1e-4 and decaying to 1e-6 via cosine annealing \cite{loshchilov2016sgdr}. We trained for 100 epochs on the mixed dataset using 8 NVIDIA A100 GPUs at a resolution of 256x256. Due to varying data volumes for different tasks, we adjusted the frequency of random patch cropping across datasets to balance data quantity. For data augmentation, we applied horizontal and vertical flips to all datasets. The model used the U-Net \cite{dhariwal2021diffusion} architecture, DDIM \cite{song2020denoising} as the sampling strategy during inference, and 4 timesteps for all tasks.

\subsection{Comparison with state-of-the-art methods}
To comprehensively evaluate the effectiveness of Defusion, we compared it with general architecture-based image restoration methods and unified frameworks. \cref{table:aio} compares performance with state-of-the-art (SOTA) methods across 8 tasks. Across all datasets, Defusion outperformed other unified methods. As a unified model, Defusion even surpasses task-specific methods in several cases.
Compared to text-based instructions \cite{luo2023controlling}, ours achieved significant improvements across all tasks. Our method also outperforms learned task embeddings \cite{ai2024multimodal,potlapalli2024promptir}, sometimes by large margins.
Defusion also achieves best perceptual qualities (see the supplementary material).

\cref{fig:aio} and \cref{fig:real} show qualitative results on various datasets. In \cref{fig:aio}, our method demonstrates superior dehazing. In the deraining task (\cref{fig:aio}), our method more effectively removes rain streaks and restores textures in occluded regions. We achieved sharper results with fewer artifacts for deblurring (\cref{fig:aio}). In the real-world datasets, We provide better de-blurring and de-snowing performance (\cref{fig:real}). For simplicity, visual results for other tasks are provided in the supplementary material. Additionally, \cref{table:mix} compares the mixed distortion dataset involving rain, snow, and noise. Our method demonstrated a significant advantage in handling challenging mixed-distortion images. \cref{table:underwater} provides results in an underwater dataset, which exemplify the superior performance of our method in the face of natural mixing distortions.

\subsection{Ablation Study}
In this section, we conduct ablation studies on visual grounding, visual instruction embedding, and degradation diffusion to demonstrate the effectiveness of the different components proposed in Defusion. Naive diffusion is used to exclude diffusion performance in degradation space. Blank grounding consists of a blank image, and simple grounding is a simple striped image. Both are used to test the high response of Grounding to distortion. Text and non-instruction are used to compare the performance of Visual Instruction instructions. The results demonstrate the excellent performance of our visual instruction and degradation diffusion designs.

\section{Conclusion}
\label{sec:conclusion}

We present \textbf{Defusion}, a novel image restoration framework utilizing visual instruction-guided degradation diffusion. Defusion aligns visual instructions with degradation patterns, surpassing traditional methods dependent on ambiguous text-based priors or task-specific models. Operating in degradation space, our model reconstructs high-quality images by denoising the shift between LQ and HQ images, enhancing stability and performance. Extensive experiments demonstrate Defusion's state-of-the-art performance in various image restoration tasks, including complex real-world scenarios with multiple degradations. The proposed approach sets new state-of-the-art for multitask image restoration with a unified model based on generative models, providing a robust solution to handle a wide range of restoration applications in the real world.

\noindent\textbf{Acknowledgments} This work was supported by Beijing Natural Science Foundation (JQ24022), CAAI-Ant Group Research Fund CAAI-MYJJ 2024-02, the National Natural Science Foundation of China (No. 62372451, No. 62192785, No. 62372082, No. 62403462), the Key Research and Development Program of Xinjiang Uyghur Autonomous Region, Grant No. 2023B03024. 
{
    \small
    \bibliographystyle{ieeenat_fullname}
    \bibliography{main}
}

\clearpage
\setcounter{page}{1}
\maketitlesupplementary

\section{Additional Details of visual instruction}

\begin{figure}[!h]
    \centering
    \includegraphics[width=\columnwidth]{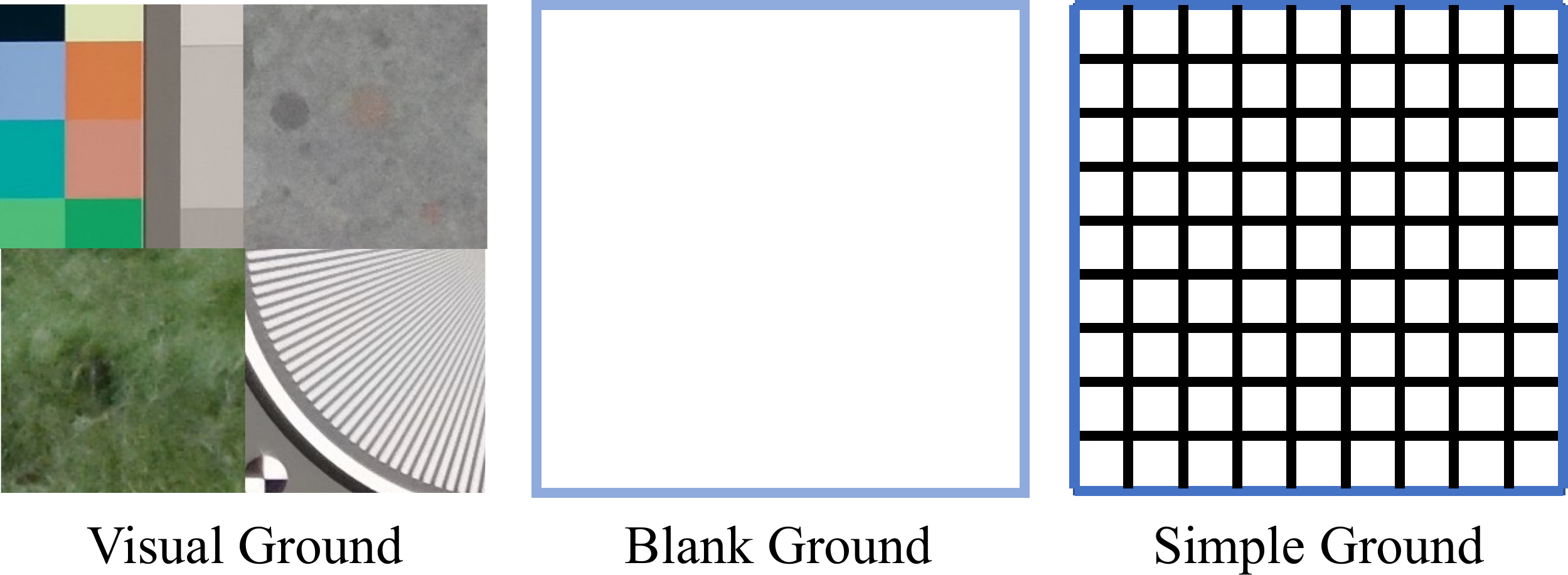}
    \caption{Replacement of visual ground.}
    \label{fig:fig_suppl3}
\end{figure}

\begin{figure}[t]
  \centering
  \vspace{-0.0cm}
  \includegraphics[width=1\linewidth]{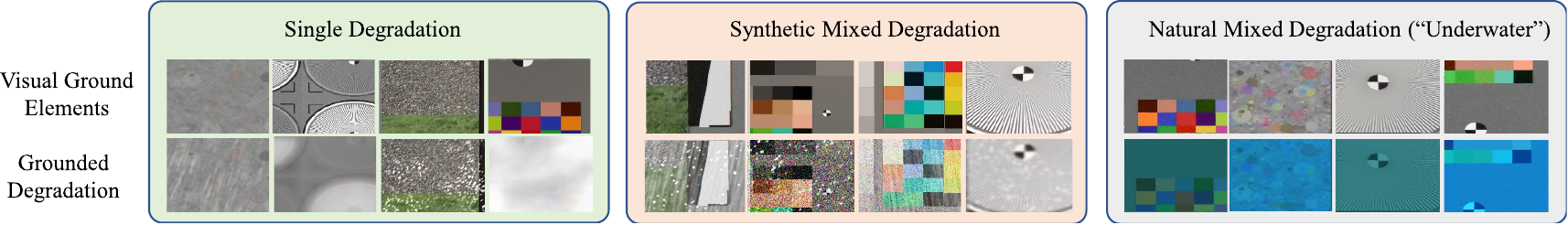}
   \label{fig:suppl_v}
   \caption{Samples of visually degraded elements from the pool of visual grounds.}
   \vspace{-0.2cm}
\end{figure}

\begin{figure*}[ht]
    \centering
    \includegraphics[width=\textwidth]{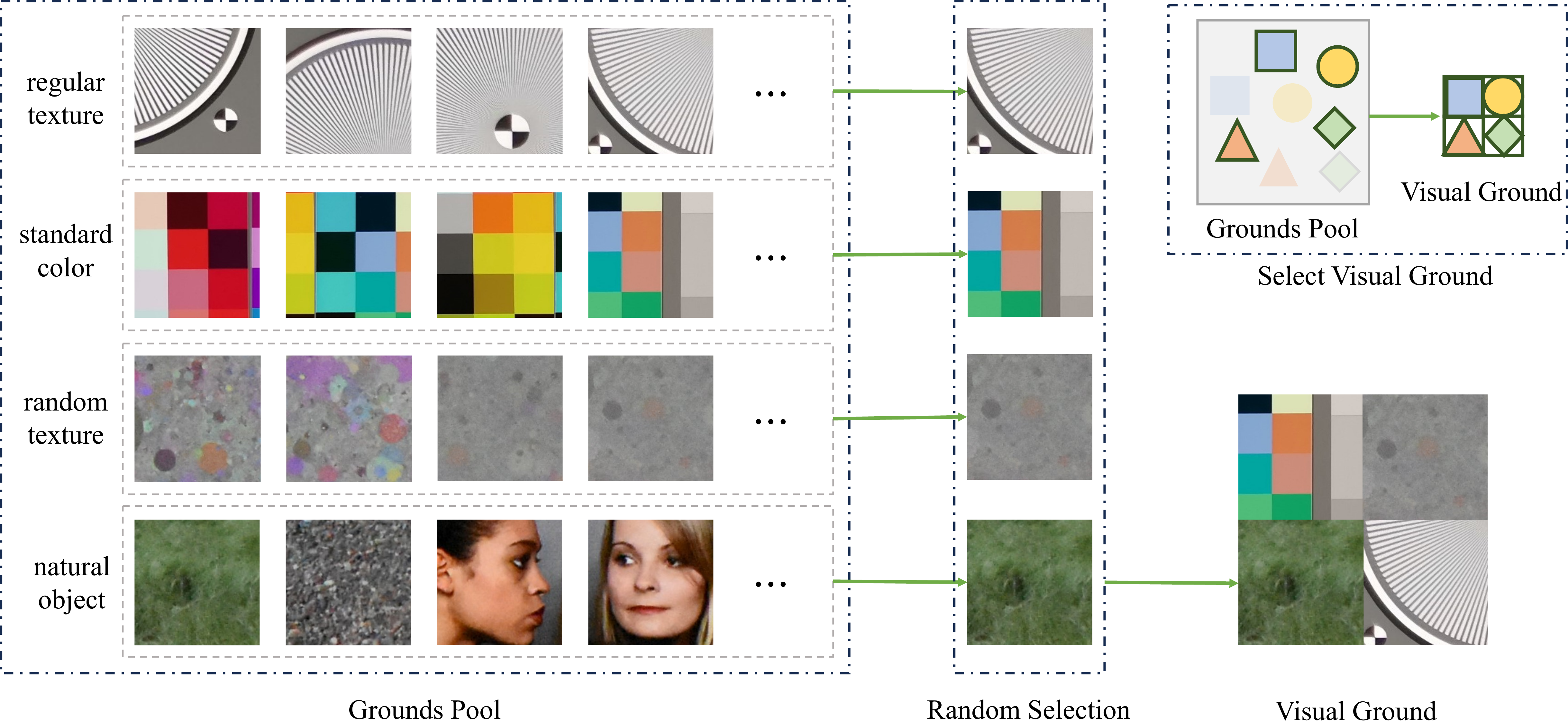}
    \caption{Details of visual ground.}
    \label{fig:fig_suppl1}
\end{figure*}

\begin{figure*}[ht]
    \centering
    \includegraphics[width=\textwidth]{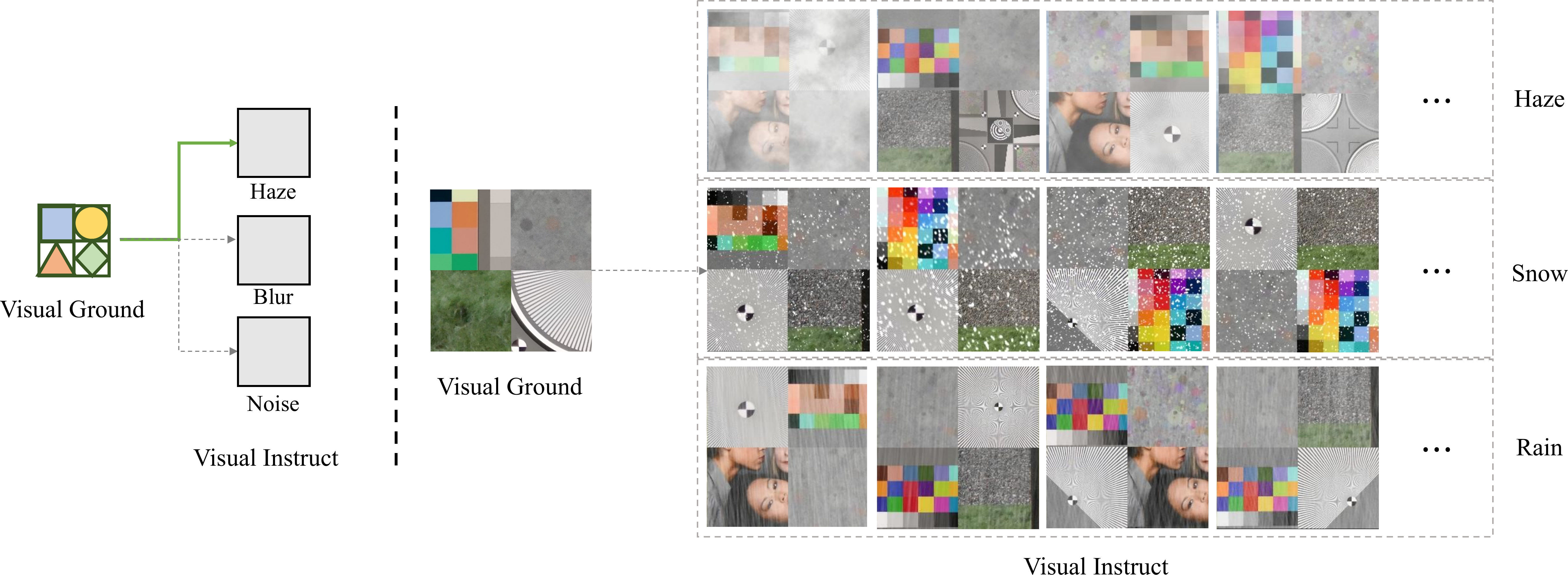}
    \caption{Samples of visual instruct.}
    \label{fig:fig_suppl2}
\end{figure*}

\begin{table*}[htp]
\caption{Comparison of perceptual metrics with state-of-the-art task-specific methods and all-in-one methods on 8 tasks. The best and second-best performances are in red and bold font, with the top 2 with a light black background.}
\renewcommand\arraystretch{1.30}
\begin{tabular}{ccccccccc}
\hline
\multicolumn{3}{c|}{\textbf{\begin{tabular}[c]{@{}c@{}}Motion Deblur\\ (GoPro \cite{nah2017deep})\end{tabular}}} & \multicolumn{3}{c|}{\textbf{\begin{tabular}[c]{@{}c@{}}Defocus Deblur\\ (DPDD \cite{abuolaim2020defocus})\end{tabular}}} & \multicolumn{3}{c}{\textbf{\begin{tabular}[c]{@{}c@{}}Desnowing\\ (Snow100K-L \cite{liu2018desnownet})\end{tabular}}} \\ \hline
\multicolumn{1}{c|}{Method} & FID$\downarrow$ & \multicolumn{1}{c|}{LPIPS$\downarrow$} & \multicolumn{1}{c|}{Method} & FID$\downarrow$ & \multicolumn{1}{c|}{LPIPS$\downarrow$} & \multicolumn{1}{c|}{Method} & FID$\downarrow$ & LPIPS$\downarrow$ \\ \hline
\textbf{Task Specific} &  &  &  &  &  &  &  &  \\ \hline
\multicolumn{1}{c|}{MPRNet\cite{zamir2021multi}} & 10.98 & \multicolumn{1}{c|}{0.091} & \multicolumn{1}{c|}{DRBNet\cite{ruan2022learning}} & 49.04 & \multicolumn{1}{c|}{0.183} & \multicolumn{1}{c|}{DesnowNet\cite{liu2018desnownet}} & - & - \\
\multicolumn{1}{c|}{Restormer\cite{zamir2022restormer}} & 10.63 & \multicolumn{1}{c|}{0.086} & \multicolumn{1}{c|}{Restormer\cite{zamir2022restormer}} & \cellcolor[HTML]{EFEFEF}{\ul \textbf{44.55}} & \multicolumn{1}{c|}{\cellcolor[HTML]{EFEFEF}{\ul \textbf{0.178}}} & \multicolumn{1}{c|}{DDMSNet\cite{zhang2021deep}} & 3.24 & 0.096 \\
\multicolumn{1}{c|}{Stripformer\cite{tsai2022stripformer}} & \cellcolor[HTML]{EFEFEF}{\ul \textbf{9.03}} & \multicolumn{1}{c|}{\cellcolor[HTML]{EFEFEF}{\ul \textbf{0.079}}} & \multicolumn{1}{c|}{NRKNet\cite{quan2023neumann}} & 55.23 & \multicolumn{1}{c|}{0.210} & \multicolumn{1}{c|}{DRT\cite{liang2022drt}} & 8.15 & 0.135 \\
\multicolumn{1}{c|}{DiffIR\cite{xia2023diffir}} & 9.65 & \multicolumn{1}{c|}{0.081} & \multicolumn{1}{c|}{FocalNet\cite{cui2023focal}} & 48.82 & \multicolumn{1}{c|}{0.210} & \multicolumn{1}{c|}{WeatherDiff\cite{ozdenizci2023restoring}} & 2.81 & 0.100 \\ \hline
\textbf{All in One} &  &  &  &  &  &  &  &  \\ \hline
\multicolumn{1}{c|}{AirNet\cite{li2022all}} & 9.65 & \multicolumn{1}{c|}{0.081} & \multicolumn{1}{c|}{AirNet\cite{li2022all}} & 58.82 & \multicolumn{1}{c|}{0.193} & \multicolumn{1}{c|}{AirNet\cite{li2022all}} & 3.92 & 0.105 \\
\multicolumn{1}{c|}{PromptIR\cite{potlapalli2024promptir}} & 15.31 & \multicolumn{1}{c|}{0.122} & \multicolumn{1}{c|}{PromptIR\cite{potlapalli2024promptir}} & 52.64 & \multicolumn{1}{c|}{0.197} & \multicolumn{1}{c|}{PromptIR\cite{potlapalli2024promptir}} & 3.79 & 0.100 \\
\multicolumn{1}{c|}{DA-CLIP\cite{luo2023controlling}} & 17.54 & \multicolumn{1}{c|}{0.131} & \multicolumn{1}{c|}{DA-CLIP\cite{luo2023controlling}} & 57.43 & \multicolumn{1}{c|}{0.201} & \multicolumn{1}{c|}{DA-CLIP\cite{luo2023controlling}} & 3.11 & 0.098 \\
\multicolumn{1}{c|}{MPerceiver\cite{ai2024multimodal}} & 10.69 & \multicolumn{1}{c|}{0.089} & \multicolumn{1}{c|}{MPerceiver\cite{ai2024multimodal}} & 48.22 & \multicolumn{1}{c|}{0.190} & \multicolumn{1}{c|}{MPerceiver\cite{ai2024multimodal}} & \cellcolor[HTML]{EFEFEF}{\ul \textbf{2.31}} & \cellcolor[HTML]{EFEFEF}{\color[HTML]{FE0000} \textbf{0.087}} \\
\multicolumn{1}{c|}{\textbf{Defusion(Ours)}} & \cellcolor[HTML]{EFEFEF}{\color[HTML]{FE0000} \textbf{8.73}} & \multicolumn{1}{c|}{\cellcolor[HTML]{EFEFEF}{\color[HTML]{FE0000} \textbf{0.052}}} & \multicolumn{1}{c|}{\textbf{Defusion(Ours)}} & \cellcolor[HTML]{EFEFEF}{\color[HTML]{FE0000} \textbf{20.20}} & \multicolumn{1}{c|}{\cellcolor[HTML]{EFEFEF}{\color[HTML]{FE0000} \textbf{0.066}}} & \multicolumn{1}{c|}{\textbf{Defusion(Ours)}} & \cellcolor[HTML]{EFEFEF}{\color[HTML]{FE0000} \textbf{0.70}} & \cellcolor[HTML]{EFEFEF}{\ul \textbf{0.094}} \\ \hline
\multicolumn{3}{c|}{\textbf{\begin{tabular}[c]{@{}c@{}}Raindrop Removal\\ (RainDrop \cite{qian2018attentive})\end{tabular}}} & \multicolumn{3}{c|}{\textbf{\begin{tabular}[c]{@{}c@{}}Deraining\\ (Rain1400 \cite{fu2017removing})\end{tabular}}} & \multicolumn{3}{c}{\textbf{\begin{tabular}[c]{@{}c@{}}Real Denoising\\ (SIDD \cite{abdelhamed2018high})\end{tabular}}} \\ \hline
\multicolumn{1}{c|}{Method} & FID$\downarrow$ & \multicolumn{1}{c|}{LPIPS$\downarrow$} & \multicolumn{1}{c|}{Method} & FID$\downarrow$ & \multicolumn{1}{c|}{LPIPS$\downarrow$} & \multicolumn{1}{c|}{Method} & FID$\downarrow$ & LPIPS$\downarrow$ \\ \hline
\textbf{Task Specific} &  &  &  &  &  &  &  &  \\ \hline
\multicolumn{1}{c|}{AttentGAN\cite{qian2018attentive}} & 33.33 & \multicolumn{1}{c|}{0.056} & \multicolumn{1}{c|}{Uformer\cite{wang2022uformer}} & 23.31 & \multicolumn{1}{c|}{0.061} & \multicolumn{1}{c|}{MPRNet\cite{zamir2021multi}} & 49.54 & 0.200 \\
\multicolumn{1}{c|}{Quanetal.\cite{quan2019deep}} & 30.56 & \multicolumn{1}{c|}{0.065} & \multicolumn{1}{c|}{Restormer\cite{zamir2022restormer}} & 20.33 & \multicolumn{1}{c|}{0.050} & \multicolumn{1}{c|}{Uformer\cite{wang2022uformer}} & 47.18 & 0.198 \\
\multicolumn{1}{c|}{IDT\cite{xiao2022image}} & 25.54 & \multicolumn{1}{c|}{0.059} & \multicolumn{1}{c|}{DRSformer\cite{chen2023learning}} & 20.06 & \multicolumn{1}{c|}{0.050} & \multicolumn{1}{c|}{Restormer\cite{zamir2022restormer}} & 47.28 & 0.195 \\
\multicolumn{1}{c|}{UDR-S$^2$\cite{chen2023sparse}} & 27.17 & \multicolumn{1}{c|}{0.064} & \multicolumn{1}{c|}{UDR-S$^2$\cite{chen2023sparse}} & 19.89 & \multicolumn{1}{c|}{0.053} & \multicolumn{1}{c|}{ART\cite{zhang2022accurate}} & 42.38 & 0.189 \\ \hline
\textbf{All in One} &  &  &  &  &  &  &  &  \\ \hline
\multicolumn{1}{c|}{AirNet\cite{li2022all}} & 33.34 & \multicolumn{1}{c|}{0.073} & \multicolumn{1}{c|}{AirNet\cite{li2022all}} & 22.38 & \multicolumn{1}{c|}{0.058} & \multicolumn{1}{c|}{AirNet\cite{li2022all}} & 51.20 & \cellcolor[HTML]{EFEFEF}{\color[HTML]{FE0000} \textbf{0.134}} \\
\multicolumn{1}{c|}{PromptIR\cite{potlapalli2024promptir}} & 35.75 & \multicolumn{1}{c|}{0.073} & \multicolumn{1}{c|}{PromptIR\cite{potlapalli2024promptir}} & 22.59 & \multicolumn{1}{c|}{0.058} & \multicolumn{1}{c|}{PromptIR\cite{potlapalli2024promptir}} & 50.52 & 0.198 \\
\multicolumn{1}{c|}{DA-CLIP\cite{luo2023controlling}} & 29.38 & \multicolumn{1}{c|}{0.078} & \multicolumn{1}{c|}{DA-CLIP\cite{luo2023controlling}} & 35.01 & \multicolumn{1}{c|}{0.116} & \multicolumn{1}{c|}{DA-CLIP\cite{luo2023controlling}} & 34.56 & 0.186 \\
\multicolumn{1}{c|}{MPerceiver\cite{ai2024multimodal}} & \cellcolor[HTML]{EFEFEF}{\ul \textbf{19.37}} & \multicolumn{1}{c|}{\cellcolor[HTML]{EFEFEF}{\ul \textbf{0.044}}} & \multicolumn{1}{c|}{MPerceiver\cite{ai2024multimodal}} & \cellcolor[HTML]{EFEFEF}{\ul \textbf{17.82}} & \multicolumn{1}{c|}{\cellcolor[HTML]{EFEFEF}{\color[HTML]{FE0000} \textbf{0.049}}} & \multicolumn{1}{c|}{MPerceiver\cite{ai2024multimodal}} & \cellcolor[HTML]{EFEFEF}{\ul \textbf{41.11}} & 0.191 \\
\multicolumn{1}{c|}{\textbf{Defusion(Ours)}} & \cellcolor[HTML]{EFEFEF}{\color[HTML]{FE0000} \textbf{10.91}} & \multicolumn{1}{c|}{\cellcolor[HTML]{EFEFEF}{\color[HTML]{FE0000} \textbf{0.039}}} & \multicolumn{1}{c|}{\textbf{Defusion(Ours)}} & \cellcolor[HTML]{EFEFEF}{\color[HTML]{FE0000} \textbf{12.93}} & \multicolumn{1}{c|}{\cellcolor[HTML]{EFEFEF}{\ul \textbf{0.057}}} & \multicolumn{1}{c|}{\textbf{Defusion(Ours)}} & \cellcolor[HTML]{EFEFEF}{\color[HTML]{FE0000} \textbf{32.77}} & \cellcolor[HTML]{EFEFEF}{\ul \textbf{0.139}} \\ \hline
\end{tabular}%
\label{table:aio_perceptual}
\end{table*}

We focus on introducing visual instructions as they are promising in aligning with visual degradations. However, image degradations are ``dangling,'' meaning that their visual effects only manifest when they exist in the context of degraded images. Therefore, we first apply degradations on some ``standard images'' to visually demonstrate the degradations to the restoration model. We call this process \textit{grounding} of degradations, the ``standard images'' are dubbed visual grounds.

The visual grounds should contain a wide range of possible visual constructs, patterns, structures, etc., that may occur in natural images to reveal the full extent of degradation. At the same time, in order to minimize the preference of the visual grounds for degradation and enhance its representation, it should not be in some fixed form. The visual grounds has been carefully designed and consists of regular textures, random textures, standard colors and natural objects. Each part was obtained by random selection from the pool, as shown in \cref{fig:fig_suppl1}. We draw inspiration from image quality assessment \cite{mittal2012making, venkatanath2015blind, mittal2012no} and select TE42 \cite{te42}, a family of charts commonly used for camera testing and visual analysis, to construct a pool of visual grounds. TE42 comprises a rich combination of textures and color palettes.

To better encode degradations themselves (rather than image semantics), we first categorize them into regular and stochastic textures, calibrated colors, and natural images, to comprehensively encode diverse distortions. This is because different visual elements exhibit varying responses to distortions (e.g., flat regions poorly represent blur).
Each category includes numerous base elements of the corresponding category sampled from TE42. During training and inference, we sample one element from each category and combine them randomly to constrict a visual ground. Then, according to the degradation we want, multiple degradations of varying intensities are randomly applied to the visual ground.
For each degraded image that needs to be recovered, the visual ground is augmented with the same category of degradation as a visual instruction, either individually or as a mixed degradation category. Some samples of visual instructions are shown in \cref{fig:fig_suppl2}.
Finally, to better encode degradations themselves (rather than image semantics), we take the residual between the visual ground and its degraded version to facilitate independence from the image content.

In the ablation experiments, we replace visual ground with blank ground and simple ground to verify the effectiveness of the proposed visual ground. The replacement Gound is shown in \cref{fig:fig_suppl3}, where blank ground is a solid color image of equal size to the visual ground, and simple ground adds simple regular shapes.
Visualizations of some visually degraded elements from the pool of visual grounds are shown in \Cref{fig:suppl_v}. Numerous distortion-sensitive elements ensure targeted responses to specific degradations. The compositional nature of visual instructions improves generalization under compound and real-world distortions. Quantitative results on direct testing are shown in \Cref{table:mix,table:underwater}. Training on extensive synthetic distortions enables generalization to real distortions.

\section{More Details About Datasets}

Our dataset in \cref{sec:exp} consists of All-in-One datasets, mixed distortion datasets, and natural mixture datasets. 

All-in-One datasets contain images from a variety of different image recovery tasks. Our method and some of the comparison methods are trained and tested uniformly on these datasets. Details of these datasets are given below:

\begin{itemize}
    \item Motion Deblur: collected from GoPro \cite{nah2017deep} dataset containing 2103 and 1111 training and testing images, RealBlur \cite{rim2020real} dataset containing 7516 and 1961 training and testing images.
    \item Defocus Deblur: collected from DPDD \cite{abuolaim2020defocus} dataset containing 350 and 76 training and testing images.
    \item Image Desnowing: collected from Snow100K \cite{liu2018desnownet} dataset containing 50000 and 50000 training and testing images, and RealSnow \cite{zhu2023learning} dataset containing 61500(crops) and 240 training and testing images.
    \item Image Dehazing: collected from RESIDE \cite{li2018benchmarking} dataset containing 12591 training images, and Dense-Haze \cite{ancuti2019dense} dataset containing 49 and 6 training and testing images.
    \item Raindrop Removal: collected from RainDrop \cite{qian2018attentive} dataset containing 861 and 307 training and testing images, and RainDS \cite{quan2021removing} dataset containing 150 and 98 training and testing images.
    \item Image Deraining: collected from Rain1400 \cite{fu2017removing} dataset containing 12600 and 1400 training and testing images, Outdoor-Rain \cite{li2019heavy} dataset containing 8100 and 900 training and testing images, and LHP \cite{guo2023sky} (only use for testing) dataset containing 300 testing images.
    \item Real Denoising: collected from SIDD \cite{abdelhamed2018high} dataset containing 288 and 32 training and testing images.
    \item JPEG Artifact removal: training dataset collected from DIV2K and FLICKR2K \cite{Agustsson_2017_CVPR_Workshops} containing 900 and 2650 images. Testing dataset collected from LIVE1 \cite{sheikh2005live} containing 29 testing images.
\end{itemize}

Mixed distortion datasets' LQ image has three different distortions: rain, snow, and noise. These distortions are superimposed on the image of WED \cite{ma2016waterloo} dataset in all 6 orders. The reason rain, snow, and noise were chosen is because they don't conflict with each other (e.g. blur and noise). The variance of the noise is 25 and the size and speed of the rain line and snowflakes are randomized. Our method and all comparison methods train 200 iterations on this dataset for a fair comparison.

Considering the mix distortion described above as a synthetic form, we added the underwater dataset as a natural mix distortion dataset and performed image restoration. Underwater dataset collected from EUVP \cite{islam2020fast} dataset containing 515 testing images, and TURBID \cite{duarte2016dataset} dataset containing 60 testing images. Both our method and the comparison method are trained on All-in-One datasets and tested directly on the underwater dataset.

Sample visualizations for each task and dataset are shown in \cref{fig:fig_suppl3} to better understand these datasets.

\begin{figure*}[ht]
    \centering
    \includegraphics[width=\textwidth]{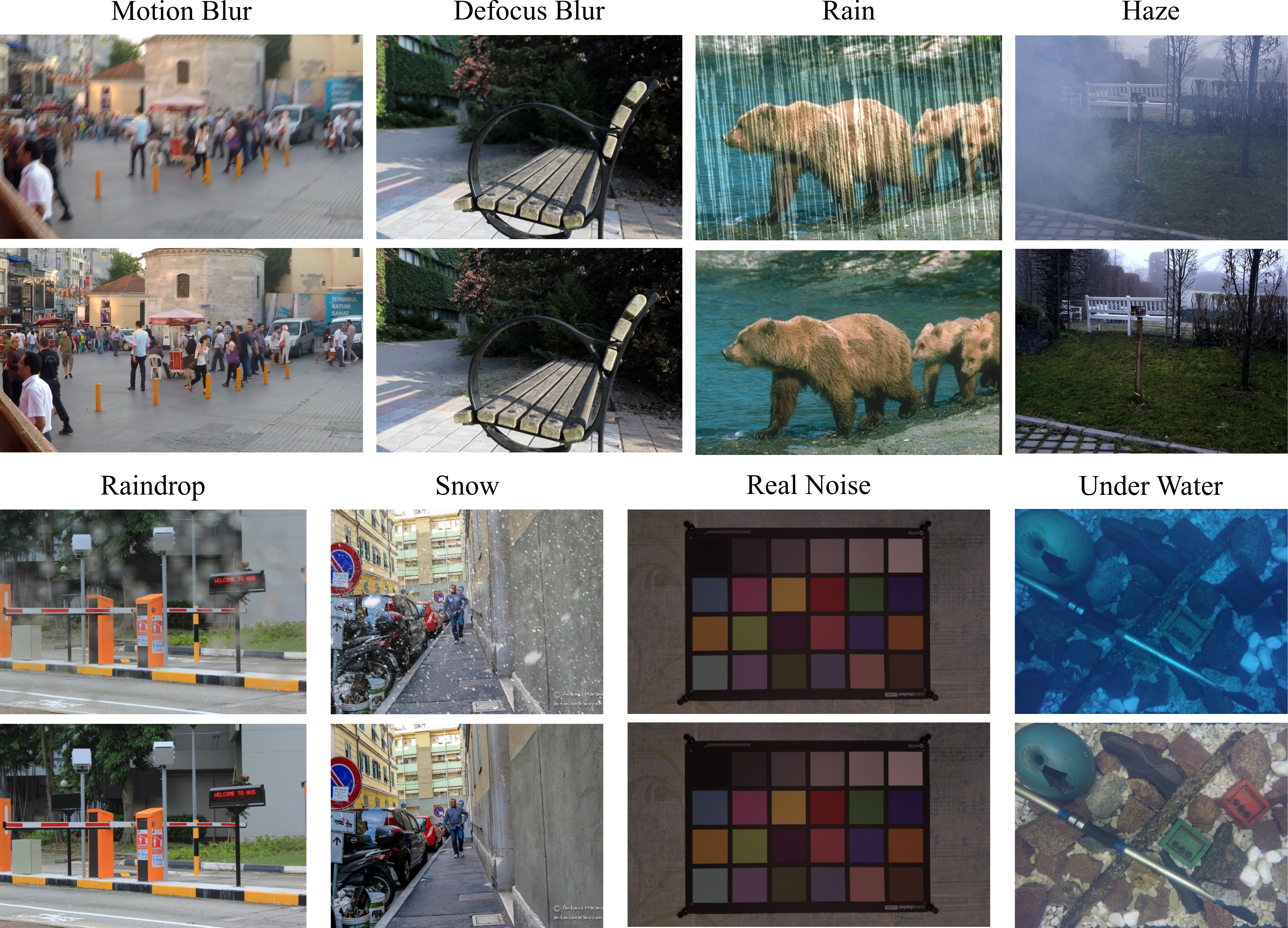}
    \caption{Samples of Datasets.}
    \label{fig:fig_suppl4}
\end{figure*}

\section{Implementation Details}

For the visual instruct tokenizer, we follow the implementation of \cite{esser2021taming}\footnote{\url{https://github.com/CompVis/taming-transformers}} and adapt its ImageNet-pretrained \cite{russakovsky2015imagenet} VQGAN model to our framework. The input size is $224\times224$, randomly cropped during training and center-cropped during inference. The embedding size is 256, and the vocabulary size is 1024. The encoder consists of five levels of channel sizes $[128, 128, 256, 256, 512]$, each level consists of two residual blocks, and the last two layers additionally have an attention block. Each level except the last one downsamples the input size by a factor of two. The decoder is symmetric to the encoder.
The visual instruct tokenizer is trained by the combination of vector quantization (VQ) loss and a reconstruction loss described by \cref{eq:loss_1,eq:loss_2,eq:loss_3}, where $\lambda=1$. Additionally, we adopt a hinge-based adversarial loss \cite{rombach2022high, esser2021taming} with a weighting of $0.8$. The discriminator follows the implementation of PatchGAN \cite{isola2017image}\footnote{\url{https://github.com/junyanz/pytorch-CycleGAN-and-pix2pix/blob/master/models/networks.py}}.
The visual instruct tokenizer is trained by the Adam optimizer \cite{kingma2014adam} with a learning rate of 4.5e-6, and $\beta_1=0.5$, $\beta_2=0.9$. The batch size is 8. We adopt random horizontal flip as data augmentation.

The base Diffusion model uses U-Net \cite{dhariwal2021diffusion} as the backbone for its restoration process, with weights pre-trained on the LAION-5B dataset \cite{schuhmann2022laion}. Built upon it, we fine-tune the Defusion on the All-in-One dataset with a batch size of 32 and with an initial learning rate of 1e-4 and decaying to 1e-6 via cosine annealing \cite{loshchilov2016sgdr}. We use the AdamW optimizer \cite{loshchilov2017decoupled} with $\beta_1$ = 0.9 and $\beta_2$ = 0.99. In preprocessing, all inputs are normalized in the range $[-1, 1]$ and randomly cropped images to $256\times 256$ for data augmentation. We train the Defusion model on eight NVIDIA A100 GPUs for a total of 600K iterations.

\section{Addition Experimental Results}

\subsection{Comparison with SOTAs on Percetual Metrics}

We also compare our method with previous SOTAs on perceptual metrics, namely FID \cite{Heusel2017GANsTB} and LPIPS \cite{Zhang2018TheUE}, which are usually more aligned with human visual preference than reference-based metrics such as PSNR and SSIM \cite{wang2004image}. The results are summarized in \Cref{table:aio_perceptual}. Note that we use the \textit{same} Defusion model and hyperparameters as for \Cref{table:aio}, while showing the best per-dataset results of the other methods. Across all datasets, Defusion achieves comparable or best perceptual qualities than both unified and task-specific methods, usually with large margins. For example, Defusion improves FID by over 100\% on defocus deblur and desnowing, while improving LPIPS by over 50\% on motion deblur and defocus deblur. On other datasets, Defusion also demonstrates advantages over previous methods. It is even more impressive considering that Defusion is optimized for reference-based metrics and generalized directly to perceptual metrics. This clearly shows the promise of diffusion-based methods with regard to human perceptual priors. We believe developing more powerful diffusion-based image restoration methods that are optimized for user preferences is a valuable future research direction.

\subsection{Visualization}

\begin{figure}[ht]
    \centering
    \includegraphics[width=\columnwidth]{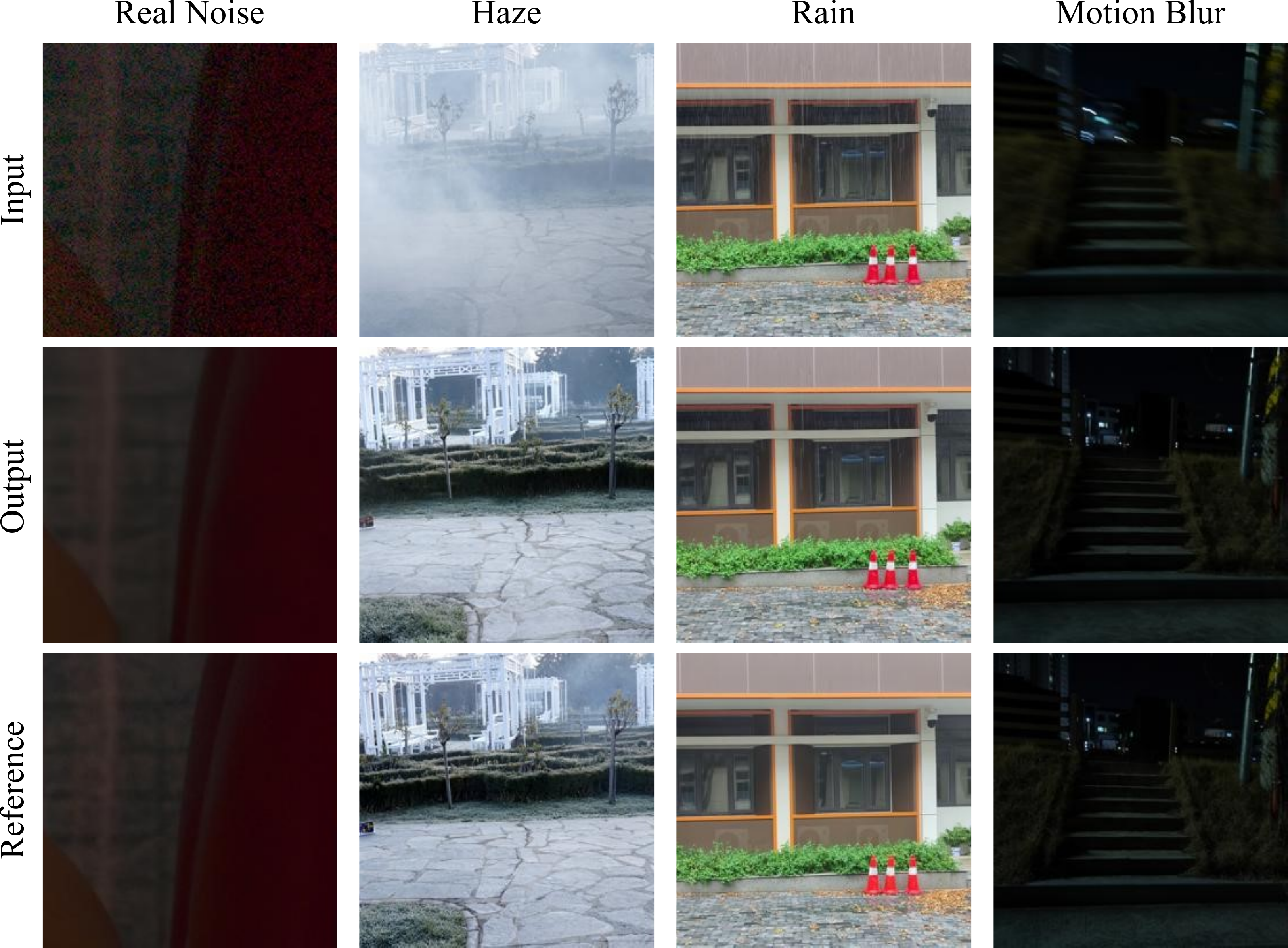}
    \caption{Visual results of real-world datasets.}
    \label{fig:fig_suppl6}
\end{figure}

\begin{figure*}[ht]
    \centering
    \includegraphics[width=\textwidth]{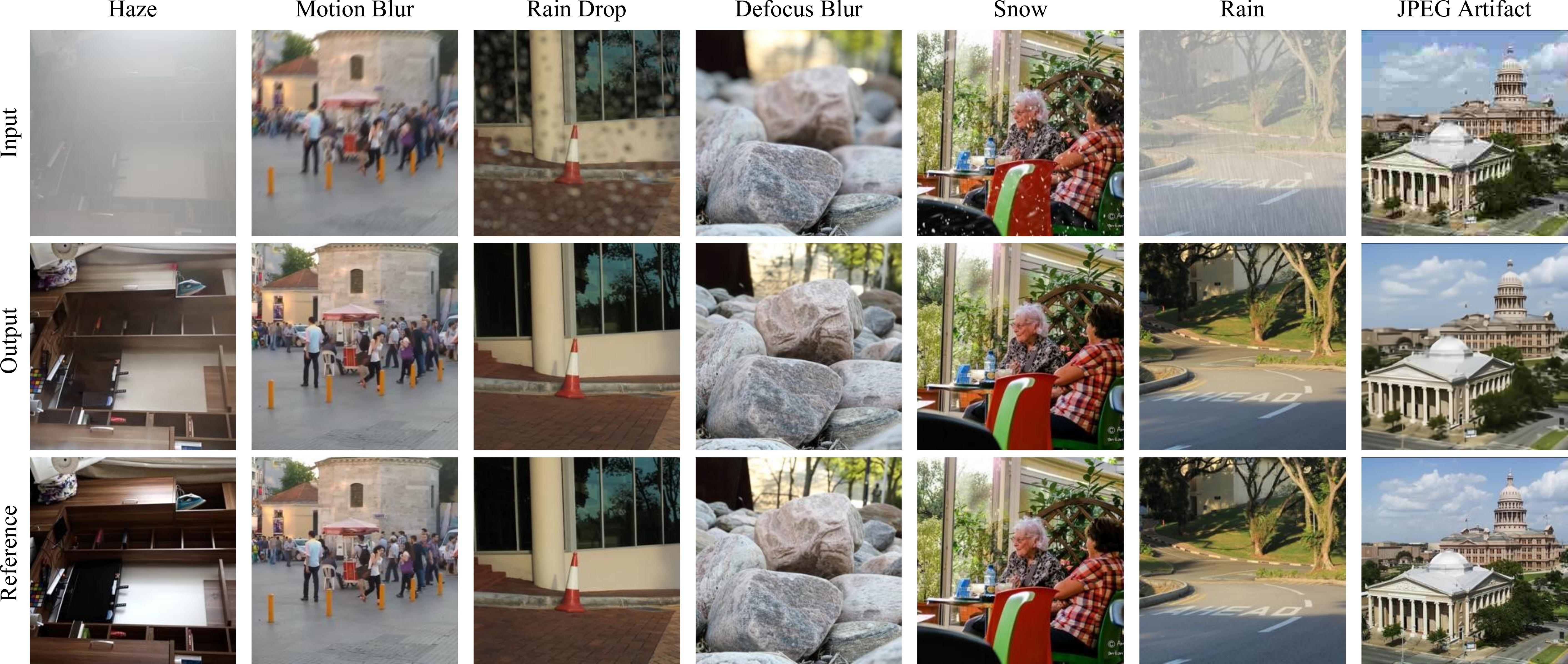}
    \caption{Visual results of synthesized datasets.}
    \label{fig:fig_suppl5}
\end{figure*}

\begin{figure*}[ht]
    \centering
    \includegraphics[width=\textwidth]{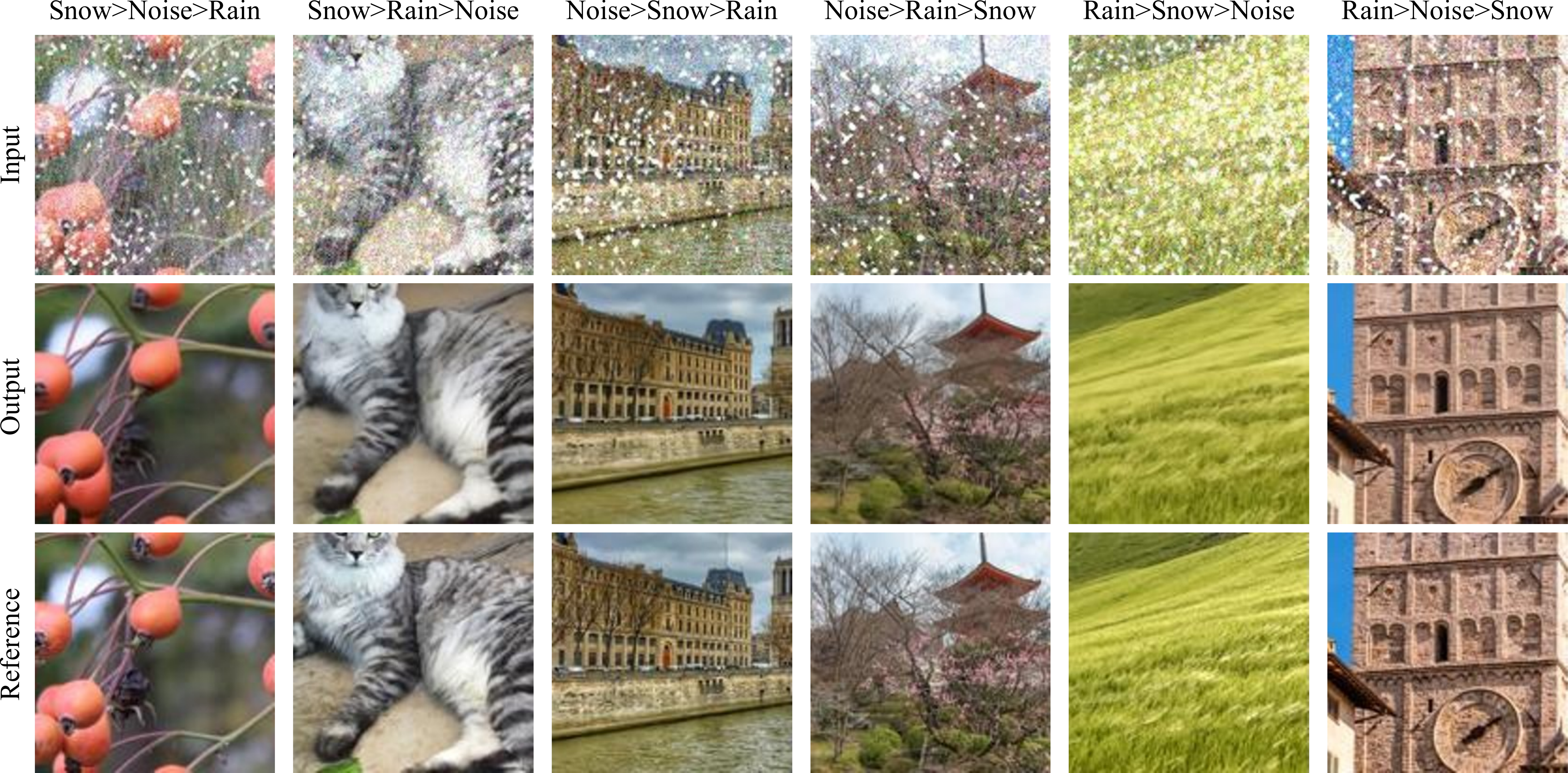}
    \caption{Visual results of mix distortion datasets.}
    \label{fig:fig_suppl7}
\end{figure*}

\Cref{fig:fig_suppl6,fig:fig_suppl5,fig:fig_suppl7} provide more visualization results on the all-in-one synthesized/real-world/mixed-degradation datasets. 

\end{document}